\documentclass[12pt]{iopart}

\usepackage{hyperref}
\usepackage{amsbsy}
\newcommand{\underset}[2]{\mathop{#2}\limits_{#1}}
\newcommand{\text}[1]{\mbox{#1}}
\newcommand{\binom}[2]{\left(\!\begin{array}{c}#1\\#2\end{array}\!\right)}
\newcommand{\eqref}[1]{(\ref{#1})}
\usepackage{tikz}
\usetikzlibrary{positioning, shapes.geometric, arrows.meta, calc}
\usepackage{algorithm}
\usepackage{algpseudocode}
\usepackage{authblk}

\begin{document}

\title[Information Filtering Networks]{Information Filtering Networks: \\ Theoretical Foundations, Generative Methodologies, and Real-World Applications}

\vspace{20pt}
\hspace{44pt} {\bf Tomaso Aste }
\address{UCL, Gower St, London WC1E 6BT, UK.  }
\ead{t.aste@ucl.ac.uk}

\vspace{10pt}
\begin{indented}
\item[]29 April 2025
\end{indented}

%

\begin{abstract} 
Information Filtering Networks (IFNs) provide a powerful framework for modeling complex systems through globally sparse yet locally dense and interpretable structures that capture multivariate dependencies. This review offers a comprehensive account of IFNs, covering their theoretical foundations, construction methodologies, and diverse applications. Tracing their origins from early network-based models to advanced formulations such as the Triangulated Maximally Filtered Graph (TMFG) and the Maximally Filtered Clique Forest (MFCF), the paper highlights how IFNs address key challenges in high-dimensional data-driven modeling. IFNs and their construction methodologies are intrinsically higher-order networks that generate simplicial complexes—structures that are only now becoming popular in the broader  literature. Applications span fields including finance, biology, psychology, and artificial intelligence, where IFNs improve interpretability, computational efficiency, and predictive performance. Special attention is given to their role in graphical modeling, where IFNs enable the estimation of sparse inverse covariance matrices with greater accuracy and scalability than traditional approaches like Graphical LASSO. Finally, the review discusses recent developments that integrate IFNs with machine learning and deep learning, underscoring their potential not only to bridge classical network theory with contemporary data-driven paradigms, but also to shape the architectures of deep learning models themselves.  
\end{abstract}


\section{Introduction}
Networks are mathematical objects composed of vertices and edges, and they serve as fundamental tools for modeling complex, interdependent systems. In these representations, vertices typically correspond to entities, variables, or features, while edges denote relationships or interactions between them \cite{newman2018networks}. While traditional network models often focus on pairwise interactions, many real-world systems exhibit dependencies that span groups of variables. These multivariate relationships are captured through higher-order structures such as cliques—fully connected subgraphs—and the connections between them \cite{battiston2020networks, bick2023higher}.

Networks enable the systematic analysis of dependencies between groups of variables that drive system dynamics, offering a powerful framework for studying systems in which the relationships among components are as critical as the components themselves. This approach has found applications across a wide range of fields—from social behavior \cite{mitchell1974social} and biological systems \cite{alm2003biological}, to economic interactions \cite{economides1996economics} and information theory \cite{shannon1948mathematical}. Representing information through a network of interactions provides crucial insights into collective behaviors and emergent properties that are otherwise difficult to discern. When integrated with quantitative models, network-based representations offer a powerful means of leveraging both individual attributes and group-level structures, revealing how information is shared, flows, and influences system-level outcomes.

Models are tools that transform data into actionable insight \cite{DDM}. Many natural and artificial systems exhibit complexity, with behavior emerging from the non-linear interactions among numerous interdependent variables \cite{holland2014complexity}. Capturing this complexity requires data-driven models that can map observed variables to outcomes, values, or categories \cite{boccara2010modeling}. These mappings may be simple, as in linear or logistic regression, or highly non-linear, as in deep learning. Regardless of complexity, the challenge remains the same: to build models that respect and exploit the intricate dependencies that structure these systems.

In response to this challenge, it was proposed in 2005 that a particularly suitable class of networks for representing complex systems would consist of structures that capture relevant local interactions while satisfying a global topological constraint—specifically, embeddability on a hyperbolic surface of a given genus \cite{aste2005complex}. Later that year, this idea was operationalized in \cite{tumminello2005} through the construction of planar graphs on the sphere, introducing a family of network representations---the Information Filtering Networks (IFNs)---that have since proven exceptionally effective for data-driven modeling in complex systems. Their success lies in their ability to combine locally dense configurations with globally sparse architectures, striking a balance between complexity and interpretability that is essential for modeling high-dimensional, multivariate systems.

IFNs are a class of networks designed specifically to represent multivariate systems. In IFNs, each vertex corresponds to a variable, and the network is constructed to represent the system’s dependency structure by linking groups of variables that share the largest common information, while adhering to constraints of sparsity and preserving global topological properties. Rather than focusing solely on pairwise interactions, IFNs emphasize the organization of variables into cliques and the connections between these cliques, resulting in a globally sparse yet locally dense structure. This allows IFNs to capture complex higher-order relationships among variables in a way that is both computationally efficient and highly interpretable.

Initially, IFN representations were developed primarily for descriptive purposes, aiming to characterize the interrelations among components within complex systems \cite{di2005interest}. Over time, their role evolved to support tasks such as clustering and variable differentiation, facilitating the identification and grouping of related elements in high-dimensional datasets \cite{pozzi2008centrality, pozzi2013spread}. A major advancement came with the introduction of chordal graphs for graphical modeling \cite{lauritzen1988,LoGo16, massara2017network}, integrating IFNs with probabilistic frameworks. 
The use of graphs for probabilistic modeling has long been studied due to their capacity to represent complex relationships in a structured and compact way. In particular, graphical models have emerged as a central paradigm, combining graph theory with multivariate probability to model dependencies among random variables \cite{lauritzen1996}. The key advantage of chordal networks in graphical modeling lies in their ability to encode the full joint probability distribution of a multivariate system while embedding conditional independence relationships directly into the network structure \cite{lauritzen1988}.  IFNs are especially well-suited to this task in the context of undirected graphical modeling, specifically Markov Random Fields (MRFs)—models that describe sets of random variables obeying the Markov property with respect to an undirected graph. This concept traces its roots to statistical physics, particularly the Sherrington–Kirkpatrick model \cite{sherrington1975solvable}, which provided a foundational lens for understanding disordered systems. In the context of IFNs, the emphasis extends beyond classical probabilistic modeling to include the systematic construction of network topologies and their use in a broad range of analytical and computational tools, thereby broadening their relevance across disciplines.

Recently, these network-based representations have expanded their reach into modern machine learning and deep learning architectures \cite{briola2023homological, wang2023homological, briola2024hlob}. By embedding IFN topologies into learning models, both computational performance and model interpretability are enhanced, effectively bridging traditional network science with cutting-edge data-driven methodologies. This integration has opened new avenues for addressing complex modeling challenges across domains where capturing high-order relationships, reducing the number of parameters, and maintaining interpretability are crucial.

The field of IFNs partially overlaps with the broader field of \textit{simplicial complexes} \cite{salnikov2018simplicial}, particularly in their common focus on capturing and representing higher-order dependencies in complex systems. Notably, IFNs such as the \textit{Planar Maximally Filtered Graph} (PMFG), the \textit{Triangulated Maximally Filtered Graph} (TMFG) \cite{massara2016}, and the \textit{Maximal Filtered Clique Forest} (MFCF) \cite{massara2017} are, by construction, simplicial complexes. These frameworks explicitly incorporate higher-order structures, the simplices, to encode multi-variable dependencies beyond pairwise interactions. As such, they are also part of the broader domain of \textit{higher-order graphs} \cite{bianconi2021}.

Beyond IFNs, in complex systems studies, there are fields such as information geometry and topological data analysis (TDA) \cite{patania2017topological} which share many guiding principles and objectives with IFNs. However, while there are strong overlaps, significant differences also exist between the domains of IFNs and these other approaches. Indeed, IFNs focus on constructing networks from data to capture the structure of complex systems and the interactions between their variables, emphasizing clarity and efficiency in representing these relationships. In contrast, information geometry examines the underlying probabilistic distributions of systems, treating them as points on geometric manifolds and employing tools like Riemannian metrics and divergence measures to analyze and optimize their properties \cite{amari2016information}. TDA, meanwhile, investigates the global structure of data by analyzing topological features such as connected components, loops, and voids through persistent homology, uncovering multi-scale and higher-order interactions within systems \cite{petri2014homological}. Together, these approaches offer a multifaceted perspective on complex systems, with IFNs excelling in constructing interpretable network representations, information geometry providing insights into continuous variable relationships, and TDA revealing structural invariants across scales.

An extensive body of literature concerns the discovery and modeling of causation with directed networks. This domain is closely related to IFNs, because, within the Wiener-Granger framework, indications of causality can be expressed as a conditional dependency between lagged variables \cite{DDM,granger1969,pearl2000}. Directed networks, such as Bayesian Networks \cite{pearl1988}, Granger causality networks \cite{basu2015network}, or Transfer Entropy networks \cite{ursino2020transfer}, explicitly model directional influences, making them particularly suitable for the study of causal inference. However, causality often requires assumptions about the temporal ordering or underlying processes driving the system, which distinguishes it from the more general approach of modeling dependencies with undirected graphs.
In this review, I concentrate on undirected IFNs, which are, somehow, more agnostic in nature. I shall therefore focus on the aim to filter and extract the most significant relationships without necessarily inferring directionality. This focus allows for a detailed examination of methods and applications specific to IFNs, without entering into the additional complexities associated with causality analysis. I leave the causality problem to future discussions.

This paper is organized as follows. Following this introduction, which has outlined the motivation and scope of the work, Section~\ref{s.foundations} presents the theoretical foundations, explaining why network representations, through composable functions and information-based grouping, can enhance model performance. 
Section~\ref{s.Origins} revisits the historical development of IFNs, from early spanning trees to modern constructions like TMFG and MFCF. 
This context sets the stage for Section~\ref{s.Overview}, which situates IFNs among other network-based approaches in quantitative modeling. 
The IFNs' generative principles and construction algorithms, including methods for bootstrapping and validation, are detailed in Section~\ref{s.GenerativeIFNalgos}. 
Building on this, Section~\ref{s.QuantitativeG} explores the integration of IFNs into statistical and machine learning models, covering applications such as sparse inverse covariance estimation, regression, feature selection, and graphical modeling. 
Section~\ref{s.Applications} illustrates the versatility of IFNs across domains like finance, biology, psychology, and AI. 
Finally, Section~\ref{s.Conclusions} reflects on the current challenges and outlines future directions, including dynamic extensions, scalability, and deeper integration with data-driven architectures.

\section{Theoretical foundations: why network representations can enhance data-driven modeling? }\label{s.foundations}
\begin{figure}
\begin{center}
\begin{tikzpicture}[node distance=1.5cm and 1.5cm, auto, thick]

    \node[circle, draw] (x1) at (0,1.5) {$x_1$};
    \node[circle, draw] (x2) at (0,0) {$x_2$};
    \node[circle, draw] (x3) at (0,-1.5) {$x_3$};
    \node[circle, draw] (x4) at (0,-3) {$x_4$};

    \node[rectangle, draw, minimum width=2cm, minimum height=0.8cm, anchor=west] (ha) at (2,0.75) {$h_a$};
    \node[rectangle, draw, minimum width=2cm, minimum height=0.8cm, anchor=west] (hb) at (2,-1) {$h_b$};
    \node[rectangle, draw, minimum width=2cm, minimum height=0.8cm, anchor=west] (hc) at (2,-3) {$h_c$};

    \node[rectangle, draw, minimum width=2.5cm, minimum height=0.8cm, anchor=west] (h) at (5,-1) {$h$};

    \draw[->] (x1.east) -- ([yshift=5pt]ha.west);
    \draw[->] (x2.east) -- ([yshift=-5pt]ha.west);

    \draw[->] (x2.east) -- ([yshift=5pt]hb.west);
    \draw[->] (x3.east) -- ([yshift=-5pt]hb.west);

    \draw[->] (x4.east) -- (hc.west);

    \draw[->] (ha.east) -- ([yshift=5pt]h.west);
    \draw[->] (hb.east) -- (h.west);
    \draw[->] (hc.east) -- ([yshift=-5pt]h.west);

    \node[below=0.cm of ha, yshift=0.cm] (formula_ha) {$h_a(x_1, x_2)$};
    \node[below=0.cm of hb, yshift=0.cm] (formula_hb) {$h_b(x_2, x_3)$};
    \node[below=0.cm of hc, yshift=0.cm] (formula_hc) {$h_c(x_4)$};

    \node[below=.0cm of h, xshift=2.7 cm] (formula_f) {$f(x_1, x_2, x_3, x_4) = h\big(h_a(x_1, x_2), h_b(x_2, x_3), h_c(x_4)\big)$};

\end{tikzpicture}
\end{center}
\vskip-.8cm
\caption{Illustration of the network representation of the composible function \(f(x_1, x_2, x_3, x_4) = h\big(h_a(x_1, x_2), h_b(x_2, x_3), h_c(x_4)\big)\). \label{f.ComposibleF} }
\end{figure}
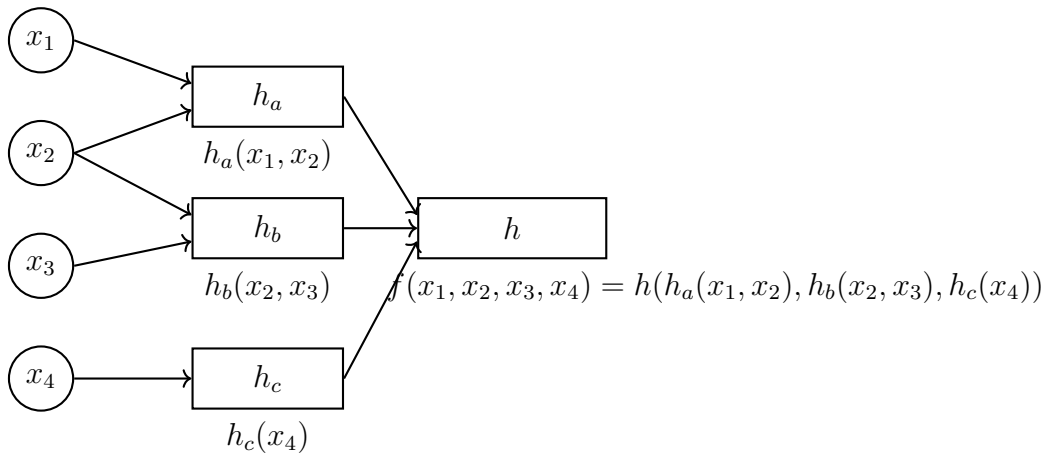

The relationship between input data and model outcomes can be represented as a function, and there is a well-established branch of mathematics, called approximation theory \cite{powell1981approximation,trefethen2019approximation}, which is dedicated to understanding how such functions can be approximated in the simplest, most manageable way. 
A key strategy for reducing the complexity of an approximate function involves grouping variables that share common information and exert similar influences on the outcome. This can be effectively achieved through the use of composite functions \cite{apostol1967one}, where a high-dimensional multivariate function (e.g., \( f(x_1, x_2, x_3, x_4) \)) is expressed as a composition of lower-dimensional functions, such as, for instance,
\begin{equation}
f(x_1, x_2, x_3, x_4) = h\big(h_a(x_1, x_2), h_b(x_2, x_3), h_c(x_4)\big),
\label{e.composableF}
\end{equation}
with each component capturing meaningful groupings of variables. This approach offers exponential reductions in complexity and addresses the curse of dimensionality by breaking a high-dimensional problem into a series of more tractable lower-dimensional subproblems, making the overall system more interpretable and computationally efficient.
It is intuitive and well-established that the hierarchical structure of composite functions can be represented with a network. In such a representation, each sub-function in the composite structure corresponds to a node, and the edges represent the dependencies between these sub-functions and their inputs or outputs. For instance, in the case of the previous function, Eq.~\ref{e.composableF}, the network would consist of nodes for each sub-function \( h_a \), \( h_b \), and \( h_c \), as well as nodes for the variables \( x_1, x_2, x_3, x_4 \), and a final aggregation node for the composite function \(h\). 
A directed edge would connect the input variables \( x_1 \) and \( x_2 \) to \( h_a \), another direct edge would connect \( x_2 \) and \( x_3 \) to \( h_b \), and a further \( x_4 \) to \( h_c \), reflecting the dependencies of these sub-functions. Similarly, the outputs of \( h_a \), \( h_b \), and \( h_c \) would connect to \( h \). This is illustrated in Fig.\ref{f.ComposibleF}.
This network representation provides a graphical and intuitive view of the composite structure, making it easier to analyze and interpret. For example, the modularity of the network allows each sub-function to be studied independently, focusing on the specific relationships between its inputs and outputs. Additionally, the hierarchical nature of the network captures the flow of information from input variables to intermediate computations and finally to the output of the composite function.
Networks of composite functions are particularly useful for simplifying complex problems \cite{poggio1990networks}. Indeed, by visualizing the function as a directed acyclic graph (DAG), one can identify clusters of nodes corresponding to groups of variables or sub-functions that share information or perform/influence related tasks. This clustering can reduce the dimensionality of optimization problems and facilitate efficient computations, as sub-functions in isolated clusters can often be processed independently or in parallel.
A prominent example of this representation is found in probabilistic graphical models where the composite function of the joint probability distribution is constructed through the aggregation of products and divisions of marginal probabilities \cite{lauritzen1996}.
More broadly, this network-based representation integrates with all machine learning frameworks. 
Neural networks, in particular, can be viewed as networks of composite functions, where each layer represents a hierarchical level of composition. The nodes in these layers correspond to activations or transformations, while the edges capture dependencies and information flow between layers, reflecting the structural and functional relationships within the network.

Grouping variables into composite components can highly simplify the discovery of good function approximators. However, determining the appropriate grouping of variables within the composite function is a nontrivial challenge. It has been argued that this discovery process can be automated during the training of deep-learning architectures. In such frameworks, it is sufficient for the hierarchical structure of the composite function to be represented as a subgraph within the network’s wiring, allowing the model to learn the optimal grouping of variables inherently. It has also been argued that this capability is a key reason why deep-learning neural networks outperform shallow networks, as their layered structure is particularly well-suited for capturing and exploiting hierarchical dependencies among variables \cite{poggio2017and}.

If the key criterion for grouping is the sharing of information among variables, it is intuitive that understanding the dependency structure would provide valuable insights into the optimal grouping for a model. IFNs are representations of the multivariate dependency structure of a system’s variables and therefore they are the best instruments to effectively inform and guide the design of  effective and accurate model architectures.
In some cases, the goodness of the model, in terms of likelihood, can be directly linked to the IFN structure. Specifically, in graphical modeling, it can be shown that the gain is equal to the mutual information between the groups of variables that are connected in the network representation. 

Identifying the network that maximally captures the shared information between variable groups is a complex problem.  As the number of variables increases, the complexity of modeling rises significantly. Specifically, interactions between couples of variables grow quadratically with the number of variables, while interactions between larger groups of variables grow at a combinatorial rate. 
Therefore, the general problem involves combinatorial optimization, with the number of possible combinations growing exponentially with the number of variables. 
This is where the methodologies developed within the IFN literature become particularly valuable, as they allow for the construction of network representations in polynomial time. The exploration of these methodologies, their effectiveness, and their applications is the primary focus of this review paper.

IFNs build on the idea that local generative moves, combined with global constraints on the overall network properties -- such as ensuring connectivity, maintaining a certain level of sparsity, or limiting the size of fully connected subgraphs (the cliques) -- provide a scalable approach to constructing large, optimized networks while avoiding the combinatorial complexity. This framework generalizes well-established algorithms like Prim's and Kruskal's algorithms \cite{prim1957shortest, newman2018networks} (see also Algorithms \ref{a.Prim} and \ref{a.Kruskal} in this paper) for constructing a minimum spanning tree. In this review, I will explore various methodologies for constructing IFNs and examine their applications, particularly in domains where IFN representations are combined with quantitative modeling. This includes perspectives from probabilistic modeling, machine learning, and deep learning, showcasing the versatility and power of IFNs in bridging network representations and data-driven analysis.

\section{Origins and Evolution of IFN Methods}\label{s.Origins}
The use of networks to represent correlation and dependency structures has deep historical roots, particularly in the context of statistical and probabilistic modeling. 
Early developments of network modeling to study correlation and causation with path diagrams were proposed by Sewall Wright’s  \cite{wright1921} with application to genetics. 
Undirected network representations first emerged within the study of covariance structures and Markov random fields. For instance, Thurstone’s work on \textit{factor analysis} \cite{thurstone1931} introduced methods for uncovering latent structures behind observed correlations, conceptually resembling undirected graphs by grouping variables into latent factors. More explicitly, Darroch, Lauritzen, and Speed formalized the use of \textit{undirected graphical models} \cite{darroch1980}, connecting Markov random fields to contingency table analysis and enabling probabilistic reasoning in high-dimensional data. In parallel, Wold’s work on \textit{partial least squares} and econometric models \cite{wold1966} contributed to dependency representation in multivariate systems. 
Lauritzen and Spiegelhalter (1988)  \cite{lauritzen1988} made a foundational contribution to the representation of dependency structures by introducing probabilistic graphical models, specifically Bayesian networks (directed) and Markov random fields (undirected). Their work provided a rigorous framework for modeling complex multivariate dependencies, enabling efficient probabilistic inference and factorization of joint distributions into simpler components. 

In Section \ref{s.GenerativeIFNalgos} I will review in detail the various generative algorithms proposed for IFNs. Let me briefly introduce them in their historical perspective. 
The first contribution to the idea of constructing a sparse network representation that maximizes dependency can be attributed to Mantegna, who proposed to use the \textit{minimum spanning tree} (MST) \cite{mantegna1999} for visualizing and characterizing the correlation structure in financial markets. This network is called \textit{minimum} because it minimizes the Euclidean distance associated with the network's edges, \( d_{i,j} = \sqrt{1-\rho_{i,j}} \), which is inversely proportional to the correlation \( \rho_{i,j} \). However, in this context, it would be more accurate to refer to the \textit{maximum spanning tree} (also, serendipitously, abbreviated as MST) since the network captures the largest dependencies (correlation coefficients in this case).
Building on the MST’s principle of capturing the relevant dependency structure through a connected network that maximizes correlations while imposing the global topological constraint of being a tree, the \textit{Planar Maximally Filtered Graph} (PMFG) extended these representations to planar graphs \cite{tumminello2005}. This allowed for a richer depiction of dependency structures while maintaining parsimony with the planarity constraints. The concept of imposing planarity as a global constraint originates from work published in the same year, where Aste, Di Matteo, and Hyde introduced an approach for embedding networks on surfaces with arbitrary genus \cite{aste2005}, with planar graphs representing the simplest class. This approach facilitates the construction and analysis of complex network ensembles that share common embeddings, providing a robust framework for investigating both local and global hierarchical properties of networks \cite{aste2012exploring}.
Later, the introduction of the \textit{Triangulated Maximally Filtered Graph} (TMFG) made two significant improvements over the PMFG \cite{massara2016}. First, it significantly enhanced computational efficiency, making it feasible to apply to much larger datasets with thousands of variables. Second, it imposed a chordal structure on the resulting network, which is crucial for certain applications, such as probabilistic graphical models and efficient inference algorithms. TMFGs are clique trees composed of tetrahedra connected through shared triangles. 
The framework was further generalized with the \textit{Maximally Filtered Clique Forest} (MFCF), which introduced a methodology based on \textit{clique expansion} to construct chordal graphs (or, equivalently, \textit{clique trees}) with cliques of arbitrary size \cite{massara2017}. This approach allows for flexible filtering of complex dependency structures while maintaining computational efficiency and ensuring compatibility with large-scale datasets \cite{DDM}. 

 In a parallel development, Kenett et al. (2010) introduced a method that combines partial correlations with PMFG to construct dependency structures by isolating direct linear relationships between variables \cite{kenett2010}. This approach seeks to effectively identify hidden relationships by filtering out indirect correlations. The use of partial correlations is particularly justified in this context because IFNs aim to represent dependencies between two variables conditioned on all others, which, in the linear case, corresponds to partial correlations.
However, estimating partial correlations becomes increasingly challenging in systems with a large number of variables due to the high-dimensional nature of the problem and the consequent curse of dimensionality. Consequently, IFNs must be better viewed as tools for approximating partial correlations rather than being constructed directly from them. I will revisit this topic in Section \ref{ss.precision}, where the estimation of the precision matrix via IFNs is discussed in detail. It must be noted that the elements of the precision matrix are the partial correlations.

\section{Brief Overview of Some Network-Based Methods that are Not IFNs}\label{s.Overview}

To provide context and draw clear distinctions, this section briefly discusses some well-known network representation methodologies that, while sharing some similarities with IFNs, developed independently and are not traditionally part of the IFN framework. Highlighting these methodologies not only underscores their individual contributions but also helps clarify their differences in purpose and construction compared to IFNs.

\subsection{Correlation Networks, Threshold Approaches, and Persistent Homology}

The construction of network representations typically begins with the challenge of selecting a relevant subgraph from the complete, weighted graph \cite{boccaletti2006}. One of the simplest and most common approaches to this problem is through thresholding. For example, a correlation matrix can be interpreted as a weighted complete graph, where the edges between nodes are the pairwise correlations between variables. A sparse subgraph can then be constructed by pruning all edges associated with correlations smaller than a given threshold value \( \rho^* \) or, similarly, that are not considered statistically significative with a p-value larger than a threshold p$^*$.  

While this thresholding method is straightforward, it suffers from significant drawbacks. Namely, it often produces either disconnected graphs or overly dense structures that fail to capture the true complexity of the system. In the context of complex systems modeling, it is crucial to retain both large dependencies and smaller, yet critical, dependencies that serve to connect different parts of the system. This distinction is where IFNs diverge from traditional correlation networks, as IFNs aim to strike a balance between sparsity and the preservation of meaningful dependencies.

In the analysis of correlation networks, the field of \textit{persistent homology} \cite{robins1999, edelsbrunner2002} has been used to study how network structures evolve as the correlation threshold varies. Persistent homology captures the birth and death of topological features, such as connected components and cycles, across different thresholds \cite{lee2012,petri2014}. However, correlation networks constructed through simple thresholding often exhibit limited topological complexity, which can diminish the utility of persistent homology in these cases \cite{DDM}. 

\subsection{Simplicial Complexes and Higher Order Networks}

Simplicial complexes extend the concept of networks by incorporating not only edges but also higher-order objects, such as triangles, tetrahedra, and their higher-dimensional analogs -- the simplexes \cite{salnikov2018simplicial}. In the literature, these network structures made of elements with a higher dimension than vertices have been named \textit{Higher-Order Networks} \cite{bianconi2021}. These structures enable the modeling of interactions beyond pairwise relationships, providing a richer framework for representing the topological and relational complexity of systems. 
Simplicial complexes are particularly valuable in contexts where higher-order interactions between groups of vertices are critical. Unlike traditional networks, which encode their structure using binary relations between vertices (the edges), simplicial complexes explicitly capture interactions between cliques. In these higher-order networks the emphasis shifts from individual relationships to group-level interactions \cite{bick2023higher}.

IFNs are simplicial complexes and higher-order networks. In fact, IFNs are an early example and practical use of higher-order networks, predating the formal coinage of the term by about a decade.  
 Subclasses of IFNs, such as the TMFG and MFCF, belong to an important category of simplicial complexes known as \textit{clique trees}. Clique trees are chordal graphs, meaning that all cycles of four or more vertices contain a chord — an edge between two non-adjacent vertices in the cycle. This chordal property ensures computational efficiency, making these structures particularly well-suited for probabilistic modeling and inference tasks.

Interestingly, most simplicial complexes discussed in the literature are also clique trees. The distinction between simplicial complexes and IFNs lies not in their structural properties but in their construction methods and their specific applications, particularly in filtering information and optimizing dependency representations.

\subsection{Bayesian Networks }
A representation of conditional independence is provided by {Bayesian Networks} (BN), which use DAGs to model conditional probability relationships. BNs are widely used, and often IFNs are confused with BN, and this is why it is important to clarify the differences first by briefly recalling the essence of BN. 
In essence, BN represent the conditional probability relation \( p(x_1, x_2) = p(x_2 | x_1) p(x_1) \), which is also an expression of Bayes' formula. In a Bayesian Network, this relation is visualized with a directed edge \( X_1 \to X_2 \), where \( X_1 \) is referred to as the \textit{parent}, and \( X_2 \) as the \textit{child}.
When \( X_1 \) has its own parent \( X_0 \), the structure \( X_0 \to X_1 \to X_2 \) represents the joint probability \( p(x_0, x_1, x_2) = p(x_2 | x_1) p(x_1 | x_0) p(x_0) \). This is a Markov chain where the Markov property assumes conditional independence between \( X_0 \) given \( X_1 \) and \( X_2 \) given \( X_1 \). In such cases, \( X_0 \) is  referred to as an \textit{ancestor}. The generalization of this structure to more complex combinations of parents, ancestors, and children within an acyclic graph defines Bayesian Networks. In this framework, a child is dependent on its parents but conditionally independent of all other ancestors.
It is important to note that the direction of an edge, such as \( X_1 \to X_2 \), does not imply conditional independence between \( X_1 \) and \( X_2 \). On the contrary, two dependent variables remain dependent in both directions. For instance, if \( p(x_2 | x_1) \neq p(x_2) \), then \( p(x_1 | x_2) \neq p(x_1) \), and the joint distribution \( p(x_1, x_2) \neq p(x_1) p(x_2) \). Consequently, parents are also conditionally dependent on their children within this framework.

In essence, BNs are directed graphs designed to characterize causality, whereas IFNs are undirected graphs used to represent dependency. While these two approaches serve distinct purposes, they can be integrated within the Wiener-Granger framework, where causality is expressed as a lagged conditional dependency measure. However, this integration falls outside the scope of this review. Readers interested in exploring these concepts further are encouraged to consult the foundational works of \cite{lauritzen1996}, particularly Section 3.2.2,  \cite{pearl2000models}, and Chapter 10 in \cite{DDM}.

\section{Construction of Information Filtering Networks  }\label{s.GenerativeIFNalgos}

Historically, IFNs have evolved over time, starting with the concept of maximum spanning trees, which focuses on capturing the most important connections in a network that connects all vertices and has the minimum number of edges. This idea expanded to the development of the Planar Maximally Filtered Graph (PMFG), which provides a more complex yet still sparse representation. Further, the Triangulated Maximally Filtered Graph (TMFG) was proposed, introducing chordal graphs. The latest development is MFCF, offering an even richer structure for representing complex systems while maintaining sparsity and chordality. In this section, I recall their construction algorithms and fundamental properties. 

\subsection{The  Maximum Spanning Tree (MST) }

The concept of the spanning tree has its origins in the work of Otakar Borůvka in 1926, who devised an efficient method for designing power grid coverage in Moravia \cite{boruuvka1926jistem,nevsetvril2001otakar}. His approach sought to connect all points requiring power with the shortest possible total distance while avoiding cycles. Such a network contains no cycles (it is a tree), which connects all vertices (it is spanning), and minimizes the total edge weight (it is minimal). Therefore, it is a \textit{Minimum Spanning Tree} (MST). Although Borůvka focused on minimizing edge distances, the complementary problem—constructing a maximum spanning tree that maximizes total edge weight is the one I focus on in this review. 

A \textit{maximum spanning tree} (MST) is a connected, undirected graph with positive edge weights that spans all vertices and has the maximum possible sum of edge weights. MSTs are subgraphs of the complete graph, containing \( p-1 \) edges for \( p = |\mathbf V| \) vertices, with no cycles and ensuring all vertices are connected by at least one path. They are particularly useful in analyzing dependency structures, where maximizing the edge weight corresponds to retaining the strongest dependencies between variables.

The construction of an MST is often achieved using Prim's or Kruskal's algorithms, which are both greedy algorithms designed to iteratively build the tree by adding edges with the highest weights while avoiding cycles. The procedure for Prim's MST construction is described in Algorithm \ref{a.Prim}.

\begin{algorithm}[H]
\caption{Prim's Algorithm for the maximum spanning tree \label{a.Prim}} 
\index{Prim's algorithm}
\begin{algorithmic}
\State {\bf Input}. A set of edges \( \mathbf E \) with positive weights \( w_{i,j} > 0 \).
\State {\bf Initialize}. Start with an empty edge set \( \mathbf E = \emptyset \) and include an arbitrary vertex \( i \) in the vertex set \( \mathbf V \):  \( \mathbf V \leftarrow \{ i \}\).
\While{\rm there are vertices not yet included in the MST, \( |\mathbf V| < p \),}
    \State Find the edge with the largest weight that connects a vertex in the MST (\( k \in \mathbf V \)) to a vertex not yet in the MST (\( j \notin \mathbf V \)):
    \begin{equation}
    (v, k) = \underset{(k,j) \in \mathbf S}{\max}{(w_{k,j} | k \in \mathbf V, j \notin \mathbf V)}.
    \end{equation}
    \State Include the vertex \( v \) in the vertex set: \( \mathbf V \leftarrow \mathbf V \cup \{v\} \).
    \State Include the edge \( (v,k) \) in the edge set: \( \mathbf E \leftarrow \mathbf E \cup \{(v,k)\} \).
\EndWhile
\State {\bf Output}. The MST: \( \mathcal G = (\mathbf V, \mathbf E) \).
\end{algorithmic}
\end{algorithm}
The procedure for Kruskal's construction is described in Algorithm \ref{a.Kruskal}.

\begin{algorithm}[H]
\caption{Kruskal's Algorithm for the Maximum Spanning Tree \label{a.Kruskal}}
\index{Kruskal's algorithm}
\begin{algorithmic}
\State \textbf{Input}. A set of edges \( \mathbf S \) with positive weights \( w_{i,j} > 0 \).
\State \textbf{Initialize}. Set \( \mathbf E = \emptyset \), \( \mathbf V = \{1, \ldots, p\} \).
\State \textbf{Initialize}. Create a forest where every vertex is a separate tree: \( \mathcal T \leftarrow \mathbf V \).
\While{there are still edges in \( \mathbf S \),}
    \State Find the edge \( (u, v) \in\mathbf S \) with the smallest weight connecting two different trees:
    \begin{equation}
    (u, v) = \underset{(k,j) \in \mathbf S}{\min}(w_{k,j} \mid k \in \mathbf t_a, j \in \mathbf t_b, \mathbf t_a \neq \mathbf t_b),
    \end{equation}
    where \( \mathbf t_a, \mathbf t_b \in \mathcal T \).
    \State Join the trees \( \mathbf t_a \) and \( \mathbf t_b \) through the new edge \( (u, v) \), creating a single tree \( \mathbf t_c \).
    \State Remove \( \mathbf t_a \) and \( \mathbf t_b \) from \( \mathcal T \) and add \( \mathbf t_c \).
    \State Remove the edge \( (u, v) \) from \( \mathbf S \) and include it in the forest's edge set: \( \mathbf E \leftarrow  \mathbf E \cup \{(u, v)\} \).
\EndWhile
\State \textbf{Output}. The MST: \( \mathcal G = (\mathbf V, \mathbf E) \).
\end{algorithmic}
\end{algorithm}
Both these algorithms are computationally efficient, with a complexity of \( \mathcal O(|\mathbf V|^2) \).
There are algorithms that can perform faster than Prim's and Kruskal's, down to almost linear time in the number of edges $\mathcal O (|\mathbf E|)$ \cite{chazelle2000minimum}. 
Despite this, Prim's and Kruskal's algorithms remain the most intuitive and widely used methods for constructing MSTs, and they are very good templates for the construction of other, more complex IFNs.

\subsection{The Planar Maximally Filtered Graph (PMFG) }

The MST problem is an optimization task that involves constructing a network with the largest possible total edge weight, subject to the constraint that the graph is connected and is a tree. Planar Maximally Filtered Graphs (PMFG) extend this concept by changing the constraint from trees to planar graphs. 

A planar graph is a graph that can be drawn on a sphere without edge crossings, and a \textit{maximal planar graph} is one to which no additional edges can be added without violating planarity \cite{whitney1933planar}. Maximal planar graphs are also called \textit{triangulations}, as they consist entirely of triangles, with \( 3p - 6 \) edges for \( p \) vertices.

The problem of finding the planar graph with the maximum edge weight, known as the Maximum Weight Planar Graph (MWPG) problem, is computationally more challenging than the MST problem, as it is NP-hard. Despite this, approximate solutions that yield suboptimal results in polynomial time have been proposed. One such method is the construction of the \textit{Planar Maximally Filtered Graph} (PMFG), proposed by \cite{tumminello2005}, which introduced the concept of IFN.

The PMFG is built using a greedy algorithm analogous to Kruskal's algorithm for the MST. The procedure iteratively adds edges with the highest weights while ensuring planarity is preserved. At the end of the process, the PMFG contains the MST as a subgraph, providing however a richer structure. The algorithm for the PMFG construction is reported in Algorithm~\ref{a.PMFG}.
\begin{algorithm}[H]
\caption{PMFG Construction for the Maximum Weight Planar Graph \label{a.PMFG}}
\begin{algorithmic}
\State \textbf{Input}. A \( p \times p \) matrix of edges with positive weights \( w_{i,j} > 0 \).
\State \textbf{Initialize}. Create an ordered set of edges in descending weight rank:
\begin{equation}
\mathbf S_k = (v_k, u_k), \text{ such that } w_{v_{k+1}, u_{k+1}} < w_{v_k, u_k}.
\end{equation}
\State \textbf{Initialize}. Start with an empty graph \( \mathbf E = \emptyset \), \( \mathbf V = \emptyset \), and set \( k \leftarrow 1 \).
\While{there are still edges to include in the PMFG (\( | \mathbf E | < 3p - 6 \))}
    \If{including edge \( \mathbf S_k = (v_k, u_k) \) does not violate planarity}
        \State Include the vertices \( v_k, u_k \) in the vertex set: 
        \begin{equation}
        \mathbf V \leftarrow \mathbf V \cup \{v_k, u_k\}.
        \end{equation}
        \State Include the edge \( (v_k, u_k) \) in the edge set: 
        \begin{equation}
        \mathbf E \leftarrow \mathbf E \cup \{(v_k, u_k)\}.
        \end{equation}
    \EndIf
    \State Increment \( k \): \( k \leftarrow k + 1 \).
\EndWhile
\State \textbf{Output}. The PMFG: \( \mathcal G = (\mathbf V, \mathbf E) \).
\end{algorithmic}
\end{algorithm}
The algorithm operates with a computational complexity of \( \mathcal{O}(p^3) \), dominated by the need to verify planarity, which is an \( \mathcal{O}(p^2) \) operation repeated \( \mathcal{O}(p) \) times during edge inclusion.

The PMFG has been widely adopted in the literature due to its ability to provide a compact and interpretable representation of dependency structures. By capturing both global and local relationships within the constraints of planarity, it offers a balance between complexity and computational feasibility, making it a valuable tool for analyzing complex systems.

An important aspect of graph embedding relates to the surface genus, which determines whether a graph can be drawn on a given surface without edge intersections. On a sphere (genus \( g = 0 \)), it is always possible to embed isolated edges and any tree, as these structures do not introduce cycles or crossings. A single cycle can also be embedded, as it is homomorphic to a circle. However, as the graph becomes more connected, embedding it on a sphere may no longer be possible. For instance, while 3-cliques and 4-cliques can be embedded on a sphere, 5-cliques cannot.  This principle is central to the {Kuratowski Theorem}, which states that a graph is planar if and only if it does not contain a subgraph that is a subdivision of \( K_5 \) or the complete bipartite graph \( K_{3,3} \) \cite{kuratowski1930probleme}.
The construction of complex networks embedded on surfaces of arbitrary genera was studied in \cite{aste2012exploring}. It is possible to embed any graph, no matter its complexity, on a surface with a large enough genus, and the relation between surface curvature and graph properties is a fascinating field of study \cite{aste2012exploring}. However, the embedding problem is NP-complete, while the generation of networks on complex surfaces is a hard-to-control mechanism. Therefore, besides the exploratory work in \cite{aste2012exploring}, to my knowledge, there have been no other successful attempts to generalize the PMFG for the construction of networks on surfaces of genera larger than zero for practical applications.

\subsection{The Triangulated Maximally Filtered Graph (TMFG) }

The \textit{Triangulated Maximally Filtered Graph} (TMFG) is a method for constructing maximally planar graphs that balance interpretability and computational efficiency. Differently from PMFG, the TMFG uses a simple \textit{clique expansion} move: adding a vertex inside a triangle on the tetrahedral simplex surface and connecting it to the triangle’s three vertices. This preserves the planarity of the graph while forming three new surface triangles \cite{massara2017network}. Repeating this process iteratively results in a maximal planar graph with \( 3p - 6 \) edges, where \( p \) is the number of vertices.

The TMFG construction was introduced to provide a computationally efficient method for filtering information in networks, particularly for large-scale systems. Unlike the PMFG, which requires verifying planarity at each step, TMFG ensures planarity by design through its construction rule (called T2 move \cite{aste2012exploring}). The algorithm for the TMFG construction is reported in Algorithm~\ref{a.TMFG}.

\begin{algorithm}[H]
\caption{TMFG Construction for the Maximum Weight Planar Graph \label{a.TMFG}}
\begin{algorithmic}
\State \textbf{Input}. A \( p \times p \) matrix of edge weights \( w_{i,j} > 0 \).
\State \textbf{Initialize}. Start with the triangle \( (u_1, u_2, u_3) \) with the largest edge weight.
\State \textbf{Initialize}. Set \( \mathbf V \leftarrow \{u_1, u_2, u_3\} \), \( \mathbf E \leftarrow \{(u_1, u_2), (u_2, u_3), (u_3, u_1)\} \), and \( \mathbf T \leftarrow \{(u_1, u_2, u_3)\} \).
\While{there are vertices not included in the TMFG (\( |\mathbf V| < p \))}
    \State Find the vertex \( v \) that maximizes the weight sum with an existing triangle \( (j_1, j_2, j_3) \) in \( \mathbf T \):
    \begin{equation}
    v = \underset{k \notin \mathbf V}{\arg\max}(w_{k,j_1} + w_{k,j_2} + w_{k,j_3}).
    \end{equation}
    \State Add three new triangles to \( \mathbf T \): \( \mathbf T \leftarrow \{(v, j_1, j_2) \), \( (v, j_2, j_3) \), \( (v, j_3, j_1)\} \).
    \State Remove \( (j_1, j_2, j_3) \) from \( \mathbf T \): \( \mathbf T \leftarrow \mathbf T \setminus \{(j_1, j_2, j_3)\}\).
    \State Update \( \mathbf V \leftarrow \mathbf V \cup \{v\} \), \( \mathbf E \leftarrow \mathbf E \cup \{(v, j_1), (v, j_2), (v, j_3)\} \).
\EndWhile
\State \textbf{Output}. The TMFG: \( \mathcal G = (\mathbf V, \mathbf E) \).
\end{algorithmic}
\end{algorithm}

The TMFG algorithm has a computational complexity of \( \mathcal O(p^2) \), which can be further optimized using a gain table that tracks the potential weight increases for each vertex-triangle pair. 
TMFG is maximally planar, ensuring it contains the maximum number of edges for a planar graph. Interested reader can access to optimized codes to generate TMFG graphs at \cite{MFCFgithub}.

\begin{figure}
\begin{minipage}{1\textwidth}
\begin{minipage}{.25\textwidth}
\begin{center}
\textbf{IFN clique tree}
\vskip.5cm
\begin{tikzpicture}[scale=1, every node/.style={circle, draw, fill=white!50, inner sep=1.5pt, font=\small}]
    \node (1) at (0,2) {1};
    \node (2) at (1,1.5) {2};
    \node (3) at (2,2) {3};
    \node (4) at (0,0) {4};
    \node (5) at (2,0) {5};
    \node (6) at (1,0.7) {6};

    \draw (1) -- (2);
    \draw (2) -- (3);
    \draw (3) -- (5);
    \draw (2) -- (4);
    \draw (2) -- (6);
    \draw (2) -- (5);
    \draw (4) -- (6);
    \draw (5) -- (6);
    \draw (4) -- (5);
\end{tikzpicture}
\end{center}
\end{minipage}
\begin{minipage}{.25\textwidth}
\begin{center}
\textbf{Hypergraph Representation}
\vskip.25cm
\begin{tikzpicture}[scale=1, every node/.style={font=\small}]

    \node[circle, draw, minimum size=1.cm] (C1) at (0,1.5) {\(1,2\)};
    \node[circle, draw, minimum size=1.2cm] (C2) at (2,0.5) {\(2,3,5\)};
    \node[circle, draw, minimum size=1.5cm] (C3) at (0,-0.5) {\(2,4,5,6\)};

    
    \draw (C1) -- (C2) node[midway, above] {\((2)\)};
    \draw (C2) -- (C3) node[midway, below] {\(\;\;(2,5)\)};
    \draw (C1) -- (C3) node[midway, left] {\((2)\)};

\end{tikzpicture}
\end{center}
\end{minipage}
\begin{minipage}{.5\textwidth}
\begin{itemize}
    \item {Number of vertices:} 6
    \item {Cliques:} \( \mathcal{C} = \{(1, 2), (2, 3, 5), (2, 4, 5, 6)\} \)
    \item {Separators:} \( \mathcal{S} = \{(2), (2, 5)\} \)
\end{itemize}
\end{minipage}
\end{minipage}
\caption{\label{f.chordalGraph}
A simple example of a chordal graph made of 6 vertices, three cliques, and two separators. 
It is a clique tree, a hypergraph with the cliques as hipervertices, and the separators as hyperedges. 
}
\end{figure}
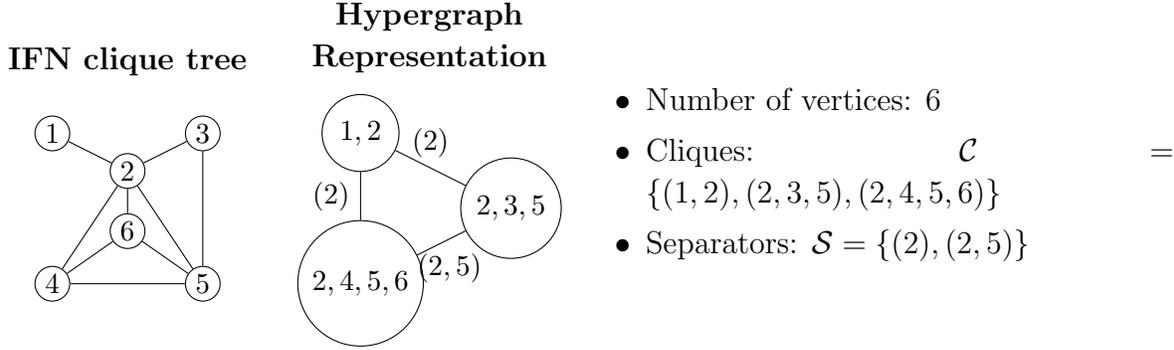

The TMFG is an example of a \textit{clique tree}, a type of graph composed of cliques connected in a tree structure by separators. A \textit{separator} is a smaller clique whose removal disconnects the graph. 
A generic example of clique tree is depicted in Fig.~\ref{f.chordalGraph}. 
In TMFGs, separators are triangles, and each triangle connects two tetrahedra. 
The TMFG structure is chordal by construction. This property ensures efficient computation and makes TMFG well-suited for probabilistic modeling tasks.
Unlike TMFG, other planar graph representations, such as the PMFG, may not be chordal and thus may not exhibit clique tree properties. 

\subsection{The Maximally Filtered Clique Forests (MFCF) }

Maximally Filtered Clique Forests (MFCFs) are clique tree constructions, where pairs of cliques (not necessarily of the same size) are joined based on a gain function that accounts for the separators shared between them. Introduced in \cite{massara2019learning}, MFCFs allow for the optimization of more complex structures beyond edge-based networks by maximizing gains associated with cliques and separators rather than individual edges. In this respect, they are natively higher-order graphs. 

The construction of an MFCF can be likened to Kruskal’s algorithm for MSTs, but it operates on cliques and separators rather than edges and vertices. The computational complexity of the problem depends on the size of the cliques involved and scales up to \( \mathcal{O}(p^{k} \log p) \), where \( p \) is the number of vertices and \( k \) is the maximum clique size. This complexity arises from the need to identify \( k \)-cliques and their shared separators, making exact solutions computationally infeasible for large systems. For practical purposes, approximate methods are used.

In analogy with the TMFG construction, MFCFs can be constructed using a \textit{clique expansion} move, where cliques are grown iteratively by attaching vertices. This approach reduces complexity significantly, producing approximate solutions in \( \mathcal{O}(p) \) once the seed clique is established. Finding a suitable seed can generally be done in \( \mathcal{O}(p^2) \). The construction algorithm is presented in Algorithm~\ref{a.MFCF}.

\begin{algorithm}[H]
\caption{MFCF Construction for the Maximal Chordal Weighted Graph  \label{a.MFCF}}
\begin{algorithmic}
\State \textbf{Input.} Gain function \( G(\cdot, \cdot) \), Min\_Cl (minimum clique size), Max\_Cl (maximum clique size), Max\_Mult (maximum separator multiplicity).
\State \textbf{Initialize.} Start with a seed clique \( \mathbf{c}_0 \) with vertices \( \mathbf{v}_0 \) and edges \( \mathbf{e}_0 \). Set \( \mathbf{V} \leftarrow \{\mathbf{v}_0\} \), \( \mathbf{E} \leftarrow \{\mathbf{e}\}_0 \), \( \mathcal{C} \leftarrow \{\mathbf{c}_0\} \).
\While{there are vertices not yet included in the MFCF (\( |\mathbf{V}| < p \)),}
    \State Find the vertex \( v \notin \mathbf{V} \) and the sub-clique \( \mathbf{s} \subseteq \mathbf{c} \in \mathcal{C} \) that maximizes the gain:
    \begin{equation}
    v = \underset{k \notin \mathbf{V}}{\arg\max}(G(k, \mathbf{s}) \mid \text{Min\_Cl} - 1 \leq |\mathbf{s}| < \text{Max\_Cl}).
    \end{equation}
    \State Create a new clique \( \mathbf{c}' = \mathbf{s} \cup \{v\} \).
    \State Add \( \mathbf{c}' \) to \( \mathcal{C} \): \(\mathcal{C} \leftarrow \mathcal{C} \cup \{\mathbf{c}'\}  \). 
    \State If \( \mathbf{s} = \mathbf{c} \), remove \( \mathbf{c} \) from \( \mathcal{C} \):  \( \mathcal{C} \leftarrow \mathcal{C} \setminus \{\mathbf{c}\} \).
    \State Update: \( \mathbf{V} \leftarrow \mathbf{V} \cup \{v\} \), \( \mathbf{E} \leftarrow \mathbf{E} \cup \{\text{all edges between \( v \) and \( \mathbf{s} \)}\} \).
\EndWhile
\State \textbf{Output.} The MFCF: \( \mathcal{G} = (\mathbf{V}, \mathbf{E}) \).
\end{algorithmic}
\end{algorithm}

The MFCF construction generalizes the TMFG by allowing arbitrary gain functions, clique sizes, and separator properties. For instance:
\begin{itemize}
    \item The MST is an MFCF with Min\_Cl = Max\_Cl = 2, where separators (vertices) can be reused multiple times and, therefore, Max\_Mult = \( p-1 \).
    \item The TMFG is an MFCF with Min\_Cl = Max\_Cl = 4, where separators (triangles) are used only once and, therefore, Max\_Mult = 1.
\end{itemize}

\begin{figure}
 \begin{center}
\centering
\includegraphics[width=0.28\textwidth]{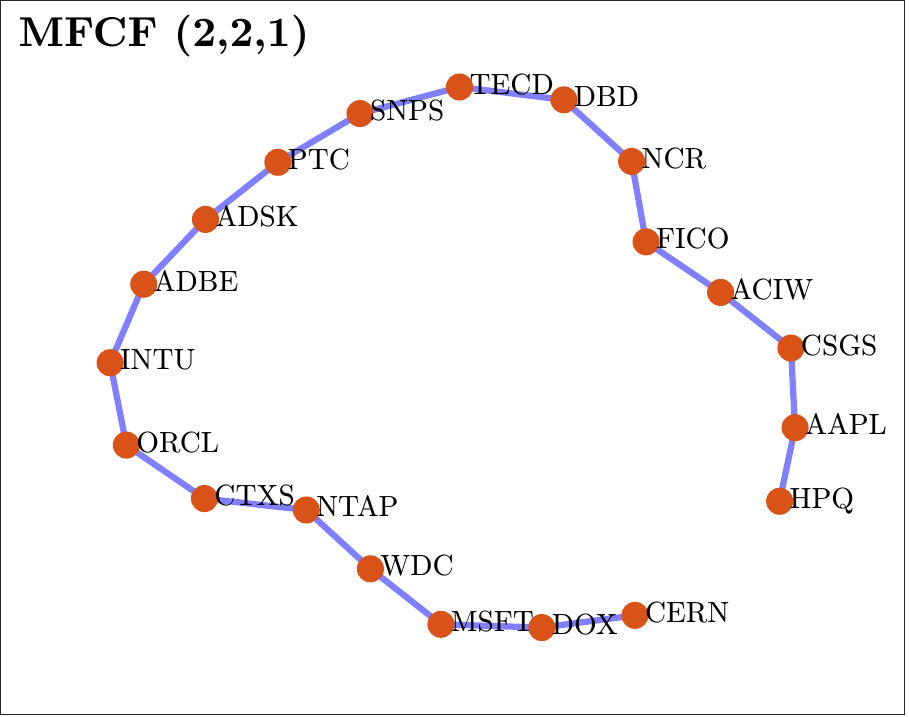}
\includegraphics[width=0.28\textwidth]{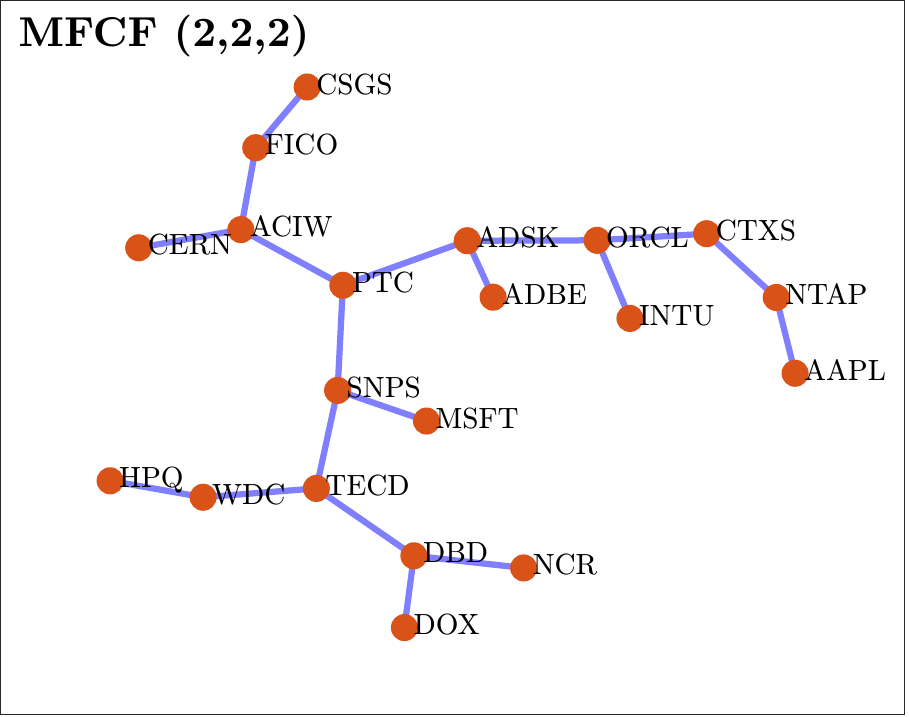}
\includegraphics[width=0.28\textwidth]{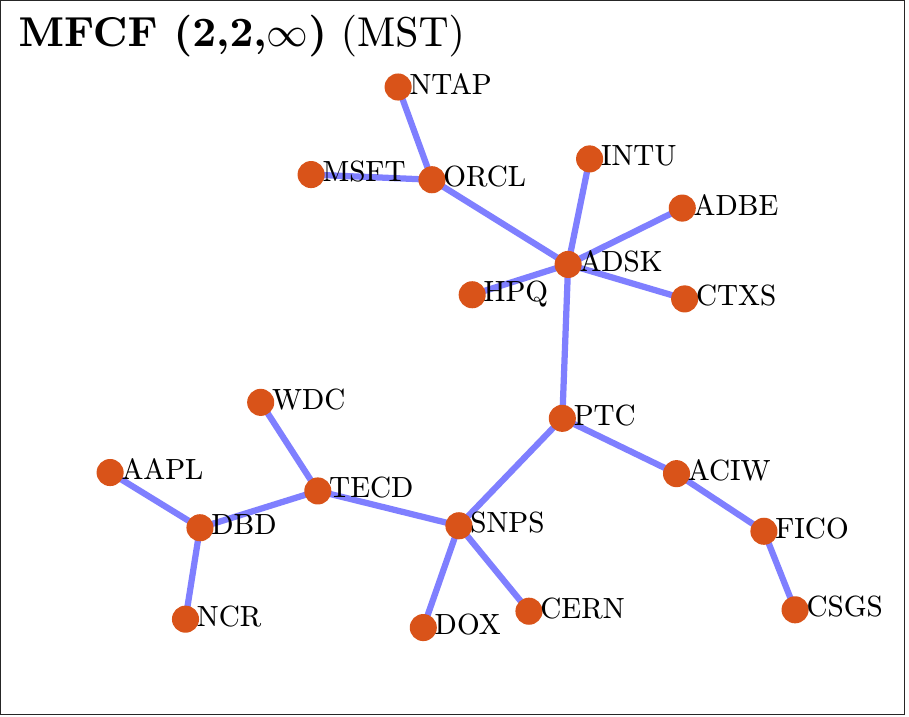}\\
\includegraphics[width=0.28\textwidth]{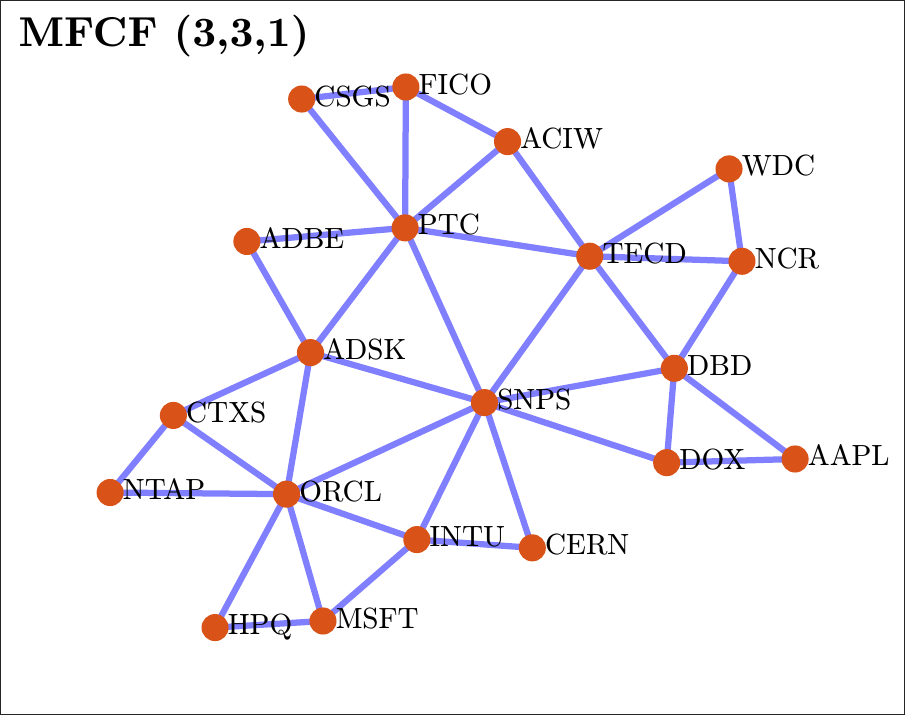}
\includegraphics[width=0.28\textwidth]{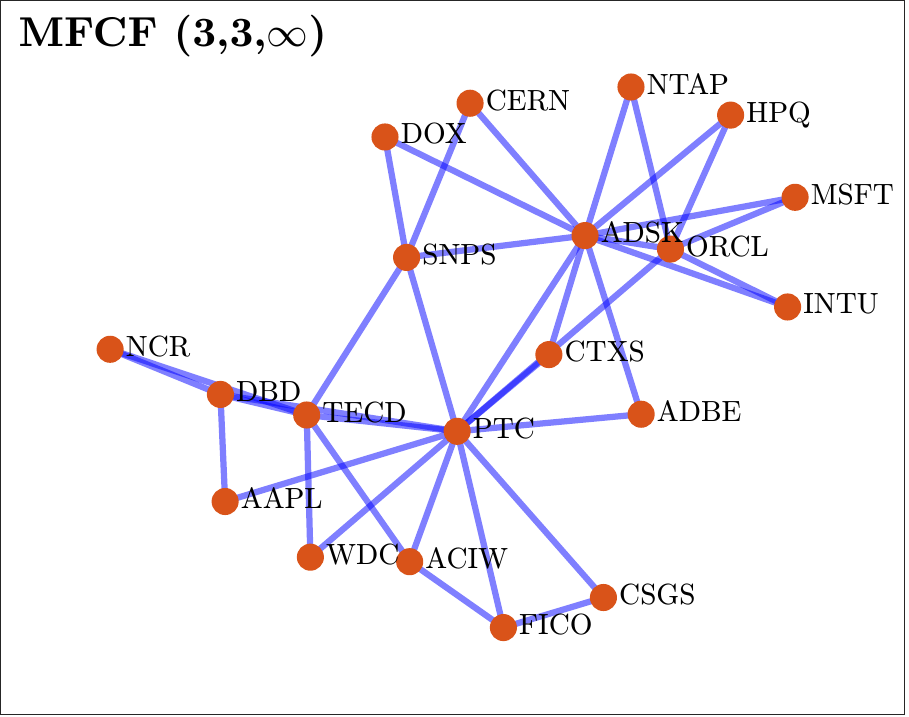}
\includegraphics[width=0.28\textwidth]{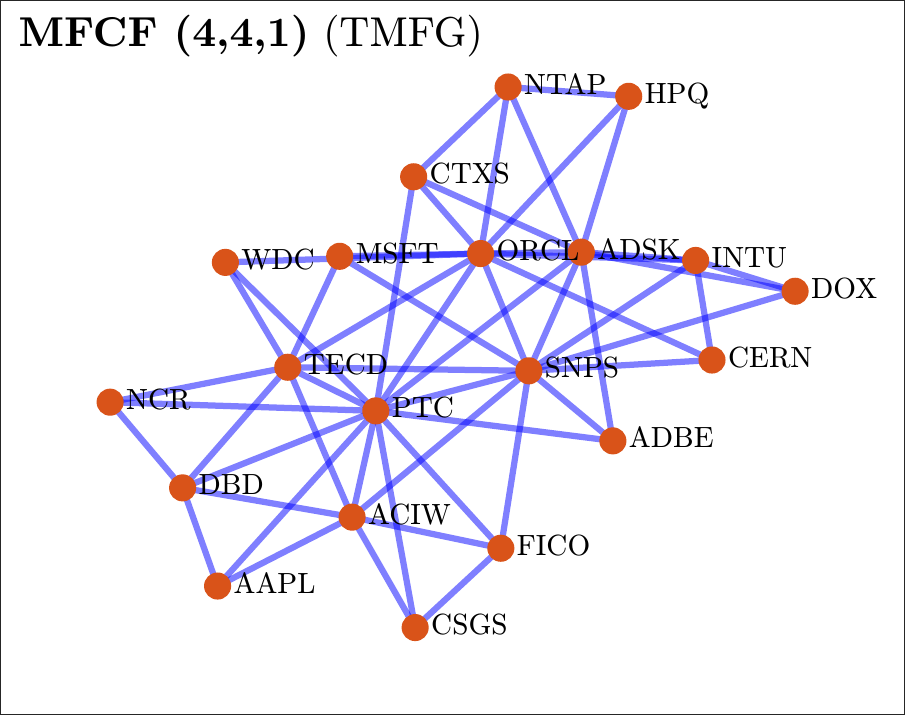}
\end{center}
\caption{\label{f.MFCF}
Examples of MFCF networks constructed to maximize the sum of correlations squared. 
The three networks on the top are trees, while the three on the bottom are planar graphs.
MFCF(2,2,1), has max and min clique sizes equal to 2, and separators can be used only once. The result can only be a line. 
MFCF(2,2,2), also has max and min clique sizes equal to 2, but vertices can have coordination up to three.
MFCF(2,2,$\infty$), is the maximum spanning tree, with max and min clique sizes equal to 2, and vertices with arbitrary coordination ($\infty$ indicates arbitrary coordination, in this case up to $p-1$).  
MFCF(3,3,1), introduces triangular cliques and separators are edges have max coordination 2.
MFCF(3,3,$\infty$), allows the separators to have arbitrary coordination. 
MFCF(4,4,1), is the TMFG (see also \cite{DDM}).
}
\end{figure}

This framework provides flexibility for constructing chordal graphs tailored to specific applications. Variations can include link validation, the ability to construct forests instead of trees, and adjustments to separator multiplicity. 
Figure \ref{f.MFCF} showcases some realizations of MFCFs. 
For further details, the reader is referred to \cite{massara2017network,DDM,massara2019learning} and the GitHub repository \cite{MFCFgithub} for implementation resources.

\subsection{Refining Network Learning by Constructing Ensembles}

The IFN construction methodologies previously described are powerful and efficient tools for generating networks that capture dependencies and possess desirable properties such as chordality or sparsity. These networks are often effective and can be directly utilized in their original form as a ``one-off'' procedure. Indeed, this approach has been predominant in the literature. However, real-world systems are inherently stochastic, and the datasets used to construct IFNs represent only a single snapshot of the underlying process. Consequently, different observations of the same system may yield slightly different network structures. 

A common method to address this variability involves bootstrapping or subsampling the dataset to generate multiple replicas. Each replica represents a perturbed version of the data, preserving key statistical properties while introducing slight variations. Constructing an IFN for each replica yields an ensemble of networks, which collectively encode the variability and uncertainty inherent in the data.

This ensemble approach opens up numerous possibilities for analyzing and refining the network structure. For instance, one can compute a probability or cumulative weight for each edge based on its frequency of appearance across the ensemble. Edges that consistently appear in many replicas are considered more ``persistent'' and likely more significant, while those that appear infrequently may represent spurious connections or noise. The resulting aggregated network can be used to extract robust structural features.

However, merging networks from the ensemble can introduce additional complexity. For example, the union of multiple chordal graphs is not guaranteed to remain chordal, potentially leading to structures that violate some of the original constraints of the IFN construction process. While this can be advantageous for exploring richer topological relationships in the data, it may also necessitate post-processing steps such as pruning or adjustments to restore properties like chordality or sparsity.

To refine the ensemble approach further, one can embed it within a Bayesian framework. By incorporating a prior distribution over network structures, such as a fully disconnected graph, and iteratively updating the posterior probability of each edge as replicas are processed.

\subsection{Network Validation}

During the IFN construction process, one aims not only to capture the largest gains (e.g., mutual information) but also to avoid including edges that are statistically insignificant or invalid. Validation is a broad topic that will not be fully addressed here; instead, let me narrow the topic to an example that is directly connected to the previous section, where the IFN is generated based on the frequencies \( f^{(r)}_{ij} \) of edge appearance across \( r \) replicas.

In this context, statistical validation can be achieved by estimating the probability that the observed number of edges between vertex $i$ and $j$, \( f^{(r)}_{ij} \), arises purely by random chance in a network with \( p(p-1)/2 \) possible placing of the edges. This probability is provided by the hypergeometric distribution, which models the likelihood of observing \( f^{(r)}_{ij} \) or more successes in \( r \) draws from a population of \( p(p-1)/2 \). The p-value for the frequency of an edge is:
\begin{equation}
P(f^{(r)}_{ij} \text{ or more}) = \sum_{k=f^{(r)}_{ij}}^{r} \frac{\binom{r}{k} \binom{p(p-1)/2 - r}{r - k}}{\binom{p(p-1)/2}{r}},
\end{equation}
quantifying the likelihood that the observed frequency is purely due to random fluctuations.

In this example of validation, edges with low values of \(P(f^{(r)}_{ij} \text{ or more})\) (the p-values) are retained as statistically significant, while those with high p-values are rejected as spurious. This validation ensures that only edges with sufficient evidence are included in the IFN. While what I have just discussed here focuses on edge-level significance, statistical validation can be performed at any level of aggregation and hierarchy. The overall network structure can also be evaluated by comparing it against random null models to ensure it reflects meaningful relationships.

\subsubsection{GitHub Repository for MFCFs Codes and Examples}  
Readers interested in exploring the implementation of IFN algorithms, including the Maximal Filtered Clique Forest (MFCF), can access the full codebase along with a wide range of application examples and datasets on the GitHub repository: \href{https://github.com/FinancialComputingUCL}{https://github.com/FinancialComputingUCL}. Contributions from the community are welcome and encouraged to further enhance the repository \cite{MFCFgithub}.

\subsection{Open Challenges for IFNs}

IFNs face several significant challenges that constrain their theoretical optimality, flexibility, scalability, and, ultimately, their applicability. Tackling these challenges has the potential to significantly expand the utility of IFNs, enabling their application to increasingly complex and large-scale problems. In this section, I explore some of the key unresolved issues. Addressing these challenges is essential for advancing the theoretical foundations of IFNs and unlocking their full potential across diverse domains.

\subsubsection{IFNs generate sub-optimal solutions} 
Among all IFNs, only the MSTs are guaranteed to achieve the theoretical optimal structure by retaining the edges with the largest (or smallest) weights necessary to maintain connectivity. This optimality can be rigorously proven by contradiction, demonstrating that, if there are no edges with the same weights, no other spanning subgraph with the same connectivity can have a higher (or lower) total weight. In contrast, the other IFNs reviewed in this section provide approximate solutions; while these methods may occasionally achieve the theoretical optimum, they lack formal guarantees. The suboptimality of these methods can be demonstrated through specific counterexamples. This limitation is intrinsic to the fact that finding subgraphs that maximize (or minimize) edge weights is generally an NP-complete problem, except in the case of trees. 

\subsubsection{Choosing the Right Gain Function}

The choice of the gain function---i.e., how much is gained by connecting two subgroups of vertices---is a flexible and critical component in the IFN construction procedure. This flexibility broadens the applicability of the method, allowing optimization problems to be tailored to specific objectives. However, quantifying the gain between groups of variables can be challenging, and this topic remains underexplored in current research.

As discussed in Section \ref{s.foundations}, the gain function should represent the shared information between the two groups of vertices. 
The mutual information  inherently quantifies higher-order relationships between variables and reflects how much information is shared between groups. However, mutual information is often computationally demanding to estimate, particularly in high-dimensional settings when more than two variables are involved, unless restrictive assumptions are imposed \cite{DDM}. 
For instance, a special case is when the INF represents the sparse structure of the inverse scale matrix of an elliptical multivariate probability. I shall discuss this in detail in Section \ref{S.EllipticalGain}.

A practical alternative that has been widely adopted is to approximate the mutual information using the sum of the squares of the pairwise correlations between all connected variables in the two groups. This proxy is computationally efficient and, as I show in Section \ref{S.EllipticalGain}, it represents the second-order approximation of mutual information under the assumption of elliptical multivariate statistics. However, this approximation inherently assumes independence within each group of variables -- a simplification that is clearly inaccurate in most cases. 
As a result, the sum-of-squares approximation fails to account for critical multi-body effects. \textit{Redundancy}, for instance, occurs when variables within the groups share overlapping information, leading to a decrease in mutual information between the two groups. Conversely, \textit{synergy} arises when the variables within the groups collectively provide more information than they do independently, increasing mutual information. These higher-order effects are essential for accurately capturing the true dependency structure between groups of variables. 

To date, systematic research into incorporating these higher-order interactions into gain functions remains limited. Addressing these aspects represents an important avenue for future work to refine the theoretical and practical foundations of IFN construction.

\subsubsection{Implications of Structural Constraints in IFN Construction }

IFNs are constructed based on predefined global constraints and specific local generative rules. These rules dictate how edges are added to the network, ensuring compliance with the structural principles underlying the chosen IFN model. Typically, IFNs aim to maximize a gain function, and in many cases, this gain increases with network density. However, the generative rules of IFNs impose constraints that limit the addition of edges once a specific structural configuration is reached (i.e. planarity or max clique size). This results in IFNs being `maximal' in the sense that no additional links can be introduced without violating the predefined generative rules.

The structural constraints introduce biases in two significant ways. The first source of bias arises from enforcing a sparse network representation, which may fail to capture dense or highly interconnected structures present in the data. The assumption that the underlying model structure is inherently sparse does not universally hold. Excluding relevant links for the sake of sparsity can be particularly problematic, as it may omit critical dependencies or interactions essential for accurate modeling.
The second source of bias stems from the strict adherence to generative rules, which may result in networks that include unnecessary edges. As noted in Section \ref{s.foundations}, the presence of unnecessary links is not inherently detrimental because training processes can often identify and disregard these redundant connections. However, while redundant links may not directly hinder the model’s ability to learn, they can still introduce challenges. Such links may mislead the training process, increase the risk of overfitting, or introduce noise, thereby complicating model convergence. This issue is particularly pronounced in high-dimensional datasets with complex and intricate dependency structures.

\subsubsection{Capturing Dynamics in Information Filtering Networks} 

This review has primarily focused on the static aspects of IFNs. However, many systems are inherently dynamic, and capturing their evolving nature requires extending IFNs into a temporal framework. Dynamical IFNs are an important area of exploration, enabling real-time or state-dependent representations of dependencies in multivariate datasets. Let me here briefly mention the challenges and a few promising ways forward. 

One of the most common methods for introducing dynamics into IFNs is through the use of rolling-window approaches. In this method, an IFN is constructed repeatedly from overlapping subsets of data, where each subset corresponds to a specific time window. This approach enables the tracking of changes in dependency structures over time. However, rolling-window methods have a significant drawback: they introduce historical inertia. Events that occurred in the past but remain within the rolling window can affect the present network structure as much as recent events, regardless of their relevance to current conditions. While this inertia can be partially mitigated by introducing a smoothing, where older observations are weighted less than more recent ones, this does not entirely eliminate the issue \cite{pozzi2012exponential}. 
The fundamental limitation lies in the reliance on historical data to infer present dependencies.
This reliance highlights a fundamental challenge in constructing real-time IFNs: the need for sufficient historical data to train models and identify meaningful dependencies. Instantaneous representations of dependency are particularly difficult to achieve in data-driven modeling, as observations require contextualization within a historical framework to extract meaningful patterns.

An alternative approach to address the limitations of rolling windows is \textit{time clustering}, which involves grouping multivariate observations into distinct temporal clusters based on a similarity measure. These clusters correspond to specific ``states'' or conditions that the system regularly revisits \cite{procacci2021forecasting}. By associating datasets and IFNs with these states, it becomes possible to construct state-dependent models that are dynamically suited to different temporal contexts. This approach offers a key advantage: it generates a set of models, each optimized for a specific state, thereby providing a more nuanced and adaptive description of the system’s dynamics. Unlike rolling-window methods, time clustering reduces the influence of historical inertia, as each IFN reflects the characteristics of a particular state rather than a continuous timeline of past observations.
With time clustering, one can recover a temporal description of the system that is better aligned with its underlying dynamics across several inherent states. This approach can capture transitions between different conditions and provide a clearer understanding of the system’s evolution over time. 
The main issue with time clustering is the identification of the relevant number of states and, more fundamentally, the assumption that these states exist and are meaningful.

\subsubsection{Adapting IFNs for Massive Datasets: Strategies and Limitations}

The scalability of IFNs to extremely large datasets presents significant challenges, particularly for cases involving millions of variables. Among IFN algorithms, the MST is notable for its computational complexity which can be reduced to \(\mathcal O(p \log p)\). However, this advantage is often overshadowed by the \(\mathcal O(p^2)\) complexity required to compute the full dependency matrix, which is a prerequisite for constructing the MST. This limitation also applies to other IFN methodologies, making scalability a critical concern. Historically, the PMFG was constrained by its \(\mathcal O(p^3)\) computational complexity, which made it unsuitable for large-scale datasets. The TMFG addressed this issue by reducing the complexity to \(\mathcal O(p^2)\). With further optimizations, such as efficient data structures, TMFG has been shown to easily handle datasets with up to \(p \sim 20,000\) variables. However, in many fields, the need to process much larger datasets, with millions of variables, persists, posing significant computational and memory challenges.

To overcome these limitations, hierarchical and modular strategies can be adopted, typically involving two main steps:

\begin{itemize}
    \item {Partitioning the Variable Space:}  
    The first step reduces the computational burden by limiting dependency calculations to smaller subsets or groups of variables. This partitioning can be achieved through clustering algorithms, domain-specific knowledge, or random sampling. 

    \item {Parallel IFN Construction and Integration:}  
    After partitioning, IFNs are constructed independently and in parallel within each subset  \cite{yu2023parallel,raphael2024faster}. The resulting partial networks are then merged to form a global IFN. \end{itemize}
This approach typically reduces the overall complexity to \(O(p \log p)\). However, this efficiency gain comes at a cost: interdependencies between variables in different subsets may be poorly represented, introducing potential biases.

While these techniques improve scalability, they present trade-offs. The selection of subsets and the merging of partial IFNs are critical steps, as they affect the accuracy and representational fidelity of the final network. Striking a balance between computational efficiency and network quality remains a central challenge.

Scalability is an essential yet underexplored area in IFN research. Advancing IFN methodologies to handle ultra-large datasets while preserving interpretability and structural integrity represents an exciting frontier. Future efforts should aim to develop scalable algorithms that maintain network accuracy while addressing the computational constraints inherent in large-scale applications.

\section{IFNs for Quantitative Modeling} \label{s.QuantitativeG}

This section delves into how IFNs can be effectively integrated with various modeling approaches, ranging from simple multilinear regressions to advanced deep-learning architectures. By combining IFNs with quantitative models, it becomes possible to develop hybrid approaches that balance interpretability, computational efficiency, and predictive accuracy.

Mathematically, the aim is to construct a model, \( \hat{f} \), in the form:
\begin{equation}
\hat{f} (\mathbf{x} | \mathcal{G}),
\end{equation}
where \( \mathbf{x} \) represents a multivariate set of variables, and \( \mathcal{G} \) denotes the IFN network that informs the model. A central challenge is identifying the optimal IFN, \( \mathcal{G} \), that provides the most meaningful structure to enhance model performance. 
The models themselves can span a variety of tasks, including regression, classification, multivariate analysis, probability estimation, or deep learning. In the following subsections, I will provide examples illustrating how IFNs can be applied in these contexts.

\subsection{IFN for Feature Selection} \label{ss.Feature}

IFNs provide a robust and efficient framework for feature selection in high-dimensional datasets in an unsupervised manner. 
This involves leveraging the IFNs topological structure to identify the relative relevance of features, enabling dimensionality reduction while retaining the most informative variables. IFNs can identify subsets of variables that preserve the essential relationships in the data while discarding redundant or irrelevant features. 

A prominent example is the Topological Feature Selection (TFS) method via IFN that was reported in \cite{briola2023topological}, where the TMFG was employed to construct a sparse network representation of the data, with nodes associated to features and edges encoding significant dependencies between them. 
Features are associated with the TMFG vertices, which are ranked based on their centrality, with highly connected nodes selected as the most relevant. 
t results that central features, often located in dense substructures or acting as hubs, are more likely to capture essential dependencies within the dataset and therefore tend to be more effective features.  
The approach is computationally efficient due to the efficiency of the TMFG and has demonstrated superior performance compared to state-of-the-art methods such as Infinite Feature Selection (Inf-FS) across a wide range of benchmark datasets.

\subsection{IFN for Covariance Selection Problem}

The covariance selection problem involves identifying significant dependencies in multivariate datasets while ensuring the resulting covariance matrix remains interpretable and computationally tractable \cite{dempster1972covariance}. IFNs provide a natural framework for addressing this challenge by focusing on the most meaningful relationships between variables, filtering out noise, and retaining a sparse but informative structure. This approach aligns with the objective of creating sparse inverse covariance matrices, which are critical for many applications, including graphical modeling and portfolio optimization.

\subsubsection{Covariance Filtering: IFN-LoGo for Sparse Inverse Covariance Estimation} \label{ss.precision}

Sparse inverse covariance estimation focuses on identifying the non-zero entries of the precision matrix (the inverse covariance matrix) associated with linear conditional independence between variables. Traditional approaches, such as Graphical Lasso (GLASSO), achieve this by employing \( \ell_1 \)-penalized likelihood methods to enforce sparsity  \cite{friedman2008sparse}. In contrast, IFNs leverage structural filtering constructing a sparse network representation that identifies the non-zero elements of the precision matrix. 

A fundamental question arises: \textit{How can the dependency structure captured by IFNs inform the conditional dependency structure required for sparse inverse covariance estimation?}

Exact inference of conditional dependencies from observed unconditional dependencies is theoretically infeasible. However, in practice, perfect inference is not required. Conditional and unconditional dependencies are intrinsically related, with conditioning either enhancing (synergy) or reducing (redundancy) interactions. This relationship ensures that, in most practical cases, the sparse structure of the IFN provides a reliable approximation of the precision matrix's structure. The key goal is to exploit dimensionality reduction and mitigate the curse of dimensionality. For this, it suffices for the exact conditional dependency structure to form a subgraph of the IFN. This is typically achieved, as IFNs are sparse graphs with locally dense structures, such as clique trees. The dense, triangulated regions around edges in an IFN make it likely that true conditional dependencies are embedded within the IFN's topology.

Once a chordal IFN is constructed (e.g. MST, TMFG, or MFCF), the sparse inverse covariance can be efficiently derived using local operations on the cliques and separators in the IFN's clique tree. 
The values of the entries of the sparse inverse covariance corresponding to an IFN edge \( (i,j) \) are computed as:
\begin{equation}\label{e.LoGo}
(\mathbf{J}^{sp})_{i,j} = \sum_{c \in \mathcal{C}} \left(\boldsymbol{\Sigma}_c^{-1}\right)_{i,j} - \sum_{s \in \mathcal{S}} \left(\boldsymbol{\Sigma}_s^{-1}\right)_{i,j},
\end{equation}
and \((\mathbf{J}^{sp})_{i,j} = 0\) if  \( (i,j) \) is not an IFN edge. 
In Eq.~\ref{e.LoGo} the sums are on the cliques and separators that contain the edge \( (i,j) \). 
The inverse covariance matrix, \(\mathbf{J}^{sp}\),  is sparse with non-zero elements coinciding with the IFN structure. This method of estimating the sparse inverse covariance is referred to as LoGo, due to its integration of \textit{local} and \textit{global} structural contributions \cite{LoGo16}. It is both computationally efficient and highly interpretable, as the computations are localized within the cliques and separators of the IFN.

The chordality of the IFN is a critical requirement for this methodology. This formula is valid for any multivariate distribution as long as the covariance is defined \cite{lauritzen1996} (see also Section \ref{ss.GM}). This highlights the robustness of IFN-LoGo as a tool for sparse inverse covariance estimation, effectively bridging structural filtering and statistical modeling.

\subsubsection{Comparison with GLASSO}

A well-known and widely celebrated method for computing sparse inverse covariance matrices is GLASSO \cite{friedman2008sparse}, which estimates the precision matrix by maximizing the penalized likelihood with an \( \ell_1 \)-norm regularization term. 
While GLASSO provides robust solutions, it requires careful tuning of the regularization parameter, which directly controls the trade-off between sparsity and accuracy. The computational cost of GLASSO can also become substantial, particularly for large datasets, when high levels of sparsity are desired or if the regularization parameter must be tuned iteratively.
Unlike GLASSO, IFN-LoGos are parameter-free and derive sparsity directly from the structural filtering process. This makes IFN-based methods computationally efficient and particularly advantageous in scenarios where interpretability, simplicity, and feasibility are critical. 

The two approaches also differ in their primary objectives. GLASSO focuses on optimizing statistical estimation by balancing data fit and sparsity through regularization, making it well-suited for dense and nuanced inference when sufficient data is available. In contrast, IFN-LoGo prioritizes structural simplicity and relevance, constructing sparse network representations that reflect the underlying relevant dependency structure of the data. This structural approach makes the IFN-LoGo approach particularly effective with limited sample sizes.

Comparisons between the two methods depend on the specific system, data characteristics, and application context. The LoGo method often excels when large sparsity is required and a small number of observations is available. Conversely, GLASSO can outperform IFNs in scenarios requiring dense or complex inference structures, provided the data volume is sufficient to support its more flexible statistical framework. Both methods can be combined with a $\ell_2$ regularization.

\subsection{IFN-LoGo for Multilinear Regressions}

In regression analyses, IFN-LoGo provides a natural framework for reducing the dimensionality of the problem while retaining essential relationships. 
The standard form of multilinear regression can be expressed as:
\begin{equation}\label{e.LinearRegerssion1}
{y} =   \mu_\mathrm{y}   + \boldsymbol \Sigma_{\mathrm{y} \mathbf x} \boldsymbol \Sigma_{\mathbf x \mathbf x}^{-1} (\mathbf x - \boldsymbol \mu_{\mathbf x}) +  \epsilon,
\end{equation}
where \( \boldsymbol \Sigma_{\mathbf x \mathbf x} \) is the covariance matrix of the predictor variables \( \mathbf x \), \( \boldsymbol \Sigma_{\mathrm{y} \mathbf x} \) is the vector of covariances between the predictors \( \mathbf x \) and the response variable \( \mathrm{y}  \),  \( \epsilon \) represents the residual error, and \(  \mu_\mathrm{y}  \) and \( \boldsymbol \mu_{\mathbf x} \) represent the expected values of \( \mathrm{y}  \) and \( \mathbf x \), respectively.

IFNs can enhance this regression framework by using the LoGo sparse expression for the inverse covariance matrix. 
Indeed, the solution to the multilinear regression problem in Eq.~\eqref{e.LinearRegerssion1} can also be expressed entirely in terms of the precision matrix \( \mathbf J = \boldsymbol \Sigma^{-1} \) of the joint system of variables \( (\mathrm{y}, \mathbf x) \), taking the form:
\begin{equation}\label{e.JmultilinearRegression}
{y} =  \mu_\mathrm{y}  - \frac{\mathbf{J}_{\mathrm{y}\mathbf x}}{\mathbf{J}_{\mathrm{yy}}} (\mathbf x - \boldsymbol \mu_{\mathbf x}) +  \epsilon,
\end{equation}
where \( \mathbf{J}_{\mathrm{y}\mathbf x} \) is the row of the precision matrix corresponding to the variable \( \mathrm{y}  \) (excluding the diagonal element), and \( \mathbf{J}_{\mathrm{yy}} \) is the diagonal element corresponding to \( \mathrm{y}  \) (see also \cite{LoGo16}). 
By using the IFN-LoGo sparse estimate of the precision matrix from Eq.~\ref{e.LoGo} into the regression Eq.~\ref{e.JmultilinearRegression}, namely
\begin{equation}
\mathbf{J} \leftarrow \mathbf{J}^{sp} ,
\end{equation}
one obtains the IFN multilinear regression. 
\begin{equation}\label{e.JmultilinearRegression1}
{y} =  \mu_\mathrm{y}  - \frac{\mathbf{J}^{sp}_{\mathrm{y}\mathbf x}}{\mathbf{J}^{sp}_{\mathrm{yy}}} (\mathbf x - \boldsymbol \mu_{\mathbf x}) +  \epsilon,
\end{equation}
This substitution has significant advantages. First, the sparse inverse covariance provides a better and lower-dimensional estimate of the precision matrix, improving numerical stability and efficiency. Second, it simplifies the regression problem by reducing it to local operations involving the variable \( \mathrm{y} \) and its first neighbors in the IFN only. By leveraging the sparse structure of IFNs, the regression is reduced to operations involving only the most relevant variables, facilitating better generalization and scalability for high-dimensional datasets.

The IFN Multilinear Regression can be seen as a specific instance of feature selection (see Section \ref{ss.Feature}), where the original regression of \( \mathrm{y}  \) with respect to the entire set of variables \( \mathbf{x} \) is reduced to a regression involving only the first neighbors of \( \mathrm{y}  \) in the IFN network. This localized approach highlights a key feature of IFNs: their ability to focus on the most relevant dependencies while significantly reducing dimensionality.

\subsection{IFN for Graphical Modeling } \label{ss.GM}

Graphical models are probabilistic frameworks that use graph structures to represent the conditional dependencies between random variables. In these models, nodes correspond to variables, while edges indicate direct conditional dependencies. Conversely, the absence of an edge between two nodes implies conditional independence between the corresponding variables. The graphical representation allows for a compact encoding of multivariate probability distributions, enabling efficient inference and learning. Notable types of graphical models include Bayesian networks, which use DAG to model causality, and Markov random fields, which rely on undirected graphs to capture mutual conditional dependencies \cite{lauritzen1996,pearl2000models}.

Graphical models can benefit significantly from IFN-based clique-tree representations. In this context, the IFN structure defines the conditional independencies between variables, facilitating efficient inference and enabling improved modeling of multivariate dependencies. A key result in this domain is the decomposition of a multivariate probability distribution \( p(\mathbf{x}) \), where \( \mathbf{x} = (x_1, x_2, \cdots, x_p) \), into a product of lower-dimensional components associated with the cliques and separators of the IFN. This decomposition is expressed as:
\begin{equation}
p(\mathbf{x})
= 
\frac{\prod_{c \in \mathcal{C}} p_c(\mathbf{x}_c)}
{\prod_{s \in \mathcal{S}} p_s(\mathbf{x}_s)},
\label{e.ProbDecomp}
\end{equation}
where \( \mathcal{C} \) and \( \mathcal{S} \) are the sets of cliques and separators in the IFN, and \( \mathbf{x}_c \) and \( \mathbf{x}_s \) are subsets of variables corresponding to these cliques and separators, respectively. This formula is a direct consequence of the Kolmogorov definition of conditional probability. Its universality makes it applicable to a broad range of distributions, provided that \( p_s(\mathbf{x}_s) > 0 \) (a condition that can be nontrivial to enforce in certain cases).

The computational efficiency and structural significance of IFNs make them highly flexible for graphical modeling. Their sparse structure reduces the complexity of inference tasks, while their computational efficiency allows for the exploration of large ensembles of potential inference structures. These ensembles can serve as Bayesian priors, which can be iteratively updated following standard Bayesian methodologies. 

If the IFN is the representation of conditional independence between variables, then the decomposition in Eq.~\ref{e.ProbDecomp} is valid for any kind of probability distribution -- including histograms. When one applies it to multivariate normal distributions, one straightforwardly obtains the LoGo expression, Eq.~\ref{e.LoGo}, for the sparse inverse covariance.

\subsection{IFN as Topological Regularizers in Probabilistic Modeling}

The use of sparse inverse covariance matrices, \( \mathbf J^{sp} \),  in parametric multivariate probability distributions, as opposed to the full inverse covariance, represents a form of topological regularization \cite{aste2020topological}. This approach restricts the model's structure to reflect only the most significant dependency relationships, effectively reducing the number of model parameters (\(\ell_0\) regularization) and simplifying the estimation process. Moreover, this topological regularization can be combined with \( \ell_2 \) regularization or shrinkage techniques targeting structured matrices, such as the constant correlation matrix. Such combinations can be efficiently applied to local estimates of the covariances within cliques and separators of the IFN. Additionally, the Expectation-Maximization (EM) algorithm can be adapted to operate on this local structure. For further implementation details, readers can refer to \cite{massara2019learning} and \cite{aste2020topological}.

A broad family of multivariate probability density functions (PDFs) that rely on the (inverse) covariance matrix belongs to the elliptical distribution family \cite{fang2018symmetric}. The PDF for this family, when defined, is expressed as:
\begin{equation}\label{e.elliptical}
f(\mathbf{x}) = k_p |\boldsymbol{\Sigma}^{-1}|^{\frac12} \mathsf{g}( d^2),
\end{equation}
where \( k_p \) is a constant, \( \mathsf{g}(\cdot) \) is the density generator function, which determines the specific member of the elliptical family. The term \( d^2 \) is a nonnegative scalar generalizing the squared Mahalanobis distance \cite{mahalanobis1936generalized}
\begin{equation}\label{e.Mahalanobis}
d = \sqrt{(\mathbf{x} - \boldsymbol{\mu})^\top \boldsymbol{\Sigma}^{-1} (\mathbf{x} - \boldsymbol{\mu})},
\end{equation}
 \( \boldsymbol{\mu} \) is the centroid of the distribution, while \( \boldsymbol\Sigma \) is a positive-definite matrix referred to as the \emph{scale matrix}. This matrix is either proportional to the covariance matrix (when defined) or identical to it, depending on the distribution. This distance provides a measure of how far \( \mathbf{x} \) is from the centroid, scaled by the metric defined by the covariance matrix.

For the elliptical multivariate probability distribution class, the mutual information between two variables $x_i$ and $x_j$ with correlation coefficient $\rho_{i,j}$ is
\begin{equation}
I(x_i;x_j) = - \frac12 \log (1-\rho_{i,j}^2) ,
\end{equation}
that, for small correlations, can be well approximated with \( I(x_i;x_j) \simeq \frac12 \rho_{i,j}^2 \).
I'll discuss the case for the mutual information between more than two variables in the next subsection (see Eq.\ref{e.MultivNorMI}). 

The multivariate normal distribution is the most widely used member of the elliptical family, characterized by \( \mathsf{g}(d^2) = \exp(-d^2/2) \). Another significant member is the multivariate Student \( t \)-distribution, with \( \mathsf{g}(d^2) = (1 + d^2/\nu)^{-(\nu + p)/2} \). For the \( t \)-distribution, when \( \nu > 2 \), the scale matrix \( \boldsymbol{\Sigma} \) is proportional to the covariance matrix, with a proportionality constant \( (\nu - 2)/\nu \).

The IFN topological regularization for this class of  PDFs is straightforward simply consisting of substituting in Eqs.\ref{e.elliptical} and \ref{e.Mahalanobis} the full inverse matrix with ist sparse IFN-LoGo  counterpart 
\begin{equation}
\boldsymbol{\Sigma}^{-1} \leftarrow \mathbf{J}^{sp} ,
\end{equation}
with \( \mathbf{J}^{sp} \)  the sparse inverse covariance computed form Eq.~\ref{e.LoGo}.
As shown in \cite{aste2020topological}, by retaining only the most important dependency relationships, this topological regularization can substantially enhance the out-of-sample performance of probabilistic models, reducing model complexity and mitigating overfitting.

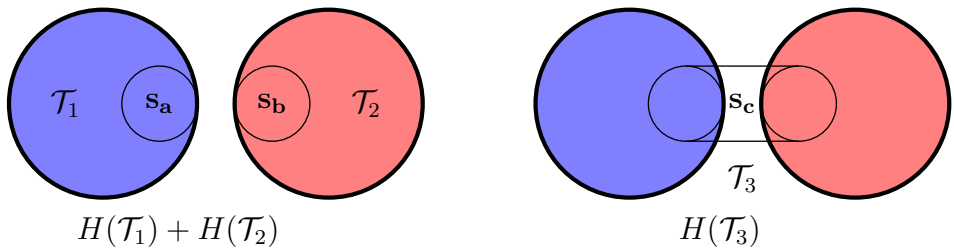
\begin{figure}
\noindent\begin{minipage}{\linewidth}\vskip10.pt
\begin{center}
\def\firstcircle{(180:1.75cm) circle (2.0cm)}
\def\secondcircle{(0:1.75cm) circle (2.0cm)}
 \begin{tikzpicture}
\tikzstyle{node_style} = [circle,draw=black,font=\rmfamily]
\tikzstyle{edge_style} = [draw=black, line width=2, ultra thick]
	\begin{scope}[shift={(2cm,-3cm)},fill opacity=0.5]
        \node [draw,
            circle,
            minimum size =2.5cm,
            fill=blue,
            font=\rmfamily,
            line width=2, 
            ultra thick] (C1) at (-7,0){};         
        \node [draw,
            circle,
            minimum size =2.5cm,
            fill=red,
            font=\rmfamily,
            line width=2, 
            ultra thick] (C2) at (-4,0){};
	   \node[fill opacity=1,font=\rmfamily] at (-7.5,0) {\(\mathcal{T}_1\)}; 
	    \node[fill opacity=1,font=\rmfamily]  at (-3.5,0) {\(\mathcal{T}_2\)};
		\node[fill opacity=1,font=\rmfamily] (v3) at (-6.25,0) {$\mathbf{{s_a}}$};
		\node[fill opacity=1,font=\rmfamily] (v4) at (-4.75,0) {$\mathbf{{s_b}}$};
		\draw[black,, line width=0.5]  (-6.25,0) circle [radius=.5cm];
		\draw[black,, line width=0.5]  (-4.75,0) circle [radius=.5cm];
	                \node[fill opacity=1,font=\rmfamily]  at (-6,-1.7) {$ H(\mathcal{T}_1)+H(\mathcal{T}_2)$};     \end{scope}
    	\begin{scope}[shift={(2cm,-3cm)},fill opacity=0.5]
        \node [draw,
            circle,
            minimum size =2.5cm,
            fill=blue,
            font=\rmfamily,
            line width=2, 
            ultra thick] (C1) at (0,0){};         
        \node [draw,
            circle,
            minimum size =2.5cm,
            fill=red,
            font=\rmfamily,
            line width=2, 
            ultra thick] (C2) at (3,0){};
	   \node[fill opacity=1,font=\rmfamily]  at (1.5,-1.) {$\mathcal{T}_3$};
		\node[fill opacity=1,font=\rmfamily] (v3) at (1.5,0) {$\mathbf{{s_c}}$};
		\draw[black,, line width=0.5]  (0.75,0) circle [radius=.5cm];
		\draw[black,, line width=0.5]  (2.25,0) circle [radius=.5cm];
	\draw[draw=black, line width=0.5]  (0.75,0.5) -- (2.25,0.5);
	\draw[draw=black, line width=0.5]  (0.75,-0.5) -- (2.25,-0.5);
	\node[fill opacity=1,font=\rmfamily]  at (1.2,-1.7) {$ H(\mathcal{T}_3)$};
    \end{scope}
 \end{tikzpicture}
\end{center}
\end{minipage}
\caption{\label{f.merge} 
A schematic representation of graphical modeling using clique tree representations. Two cliques from separate clique trees \(\mathcal{T}_1\) and \(\mathcal{T}_2\) are connected by merging the separators \(\mathbf{s}_a\) and \(\mathbf{s}_b\) into a new separator \(\mathbf{s}_c\). The resulting global gain in mutual information is given by \( I(\mathbf{s}_a ; \mathbf{s}_b) = H(\mathcal{T}_1) + H(\mathcal{T}_2) - H(\mathcal{T}_3) \). (See also \cite{DDM}.)}
\end{figure}

\subsubsection{ Higher-Order Gain Functions for Elliptical Probabilistic Modeling } \label{S.EllipticalGain}
When the IFN is a representation of the structure of the sparse inverse covariance of a multivariate elliptical distribution, there is a natural, direct, and efficient way to compute the gain function for the IFN construction in a clean and direct higher-order setting. 
Let's consider a step in the construction of the IFN when two cliques \(\mathcal{C}_a\) and \(\mathcal{C}_b\) belonging to two separated clique trees  \(\mathcal{T}_1\) and \(\mathcal{T}_2\) are joined by connecting two subsets of their vertices \( \mathbf{s}_a \in \mathcal{C}_a \), \( \mathbf{s}_b \in \mathcal{C}_b \) forming the separator  \( \mathbf{s}_c =(\mathbf{s}_a,\mathbf{s}_b) \) and the new merged clique tree  \(\mathcal{T}_3\) (see Fig.~\ref{f.merge}). 
Before the connection, the entropy associated to the multivariate system represented by the two separated clique trees is  \(H(\mathcal{T}_1) + H(\mathcal{T}_2)\). 
After the connection the entropy becomes \( H(\mathcal{T}_3) \). 
By construction, the difference between these two entropies is the mutual information of the two groups of variables forming the separator clique: 
\(H(\mathcal{T}_1) + H(\mathcal{T}_2) - H(\mathcal{T}_3) = I(\mathbf{s}_a ; \mathbf{s}_b)\). 
For multivariate elliptical distributions, such mutual information is 
 \begin{equation}\label{e.MultivNorMI}
I(\mathbf a; \mathbf  b) = \frac{1}{2} \log \frac{|\mathbf{\Sigma}_{\mathbf a}||\mathbf{\Sigma}_{\mathbf b}|}{| \mathbf{\Sigma}_{\mathbf {c}}|}, 
\end{equation} 
where \(|\cdot|\) represnt the determinat. This is a very simple expression that extends the construction of IFNs beyond pairwise interactions.

\subsection{IFN for Graphical Neural Networks} \label{ss.GNN}

Graph Neural Networks (GNNs) are a powerful class of deep learning models designed to integrate and process graph-structured data by aggregating information across nodes and edges. This approach enables GNNs to represent structured dependencies effectively. In GNNs, each node is initialized with a feature vector derived from raw data or predefined embeddings. Subsequently, information is propagated through the graph using message-passing algorithms, where a node’s representation is updated based on a weighted aggregation of its neighbors’ features. The resulting node representations are then used as input features for downstream modeling tasks.

Although IFNs and GNNs stem from distinct domains, they share similarities and can be effectively combined. 
In GNNs, the graph structure is used to locally aggregate information and refine node embeddings based on network connectivity. In IFNs, connected vertices represent input features with some degree of similarity or shared information. This makes IFNs valuable for informing and initializing GNNs by providing a meaningful topology that guides the learning of embedded representations.

The integration of IFNs into GNN frameworks has been explored with notable success, demonstrating improvements in both model performance and interpretability.
Indeed, the performance of GNNs heavily depends on the quality of the underlying graph topology, which must represent a structure in which it is meaningful to share information between neighbouring features. However, predefined or learned adjacency matrices often struggle to capture complex, high-dimensional dependencies accurately, especially in noisy or large datasets. This is where IFNs provide significant advantages.

In models like Graph Attention Networks (GAT), IFN-generated sparse graphs offer a clear prior for determining which nodes should be attended to, improving both accuracy and training stability. Additionally, since IFNs explicitly construct vertices to correspond to features and edges to represent statistically meaningful interrelations, they provide a more interpretable network structure compared to many GNN constructions. 
It has been demonstrated in \cite{wang2022network}, that the use of IFN-generated topologies in spatial-temporal GNNs can result in superior performance in time-series forecasting tasks. 

\subsection{IFN for Novel Neural Network Architectures: Homological Neural Networks (HNN)} \label{ss.HNN}

The combination of GNNs and IFNs represents a promising avenue for applying IFNs to deep learning. Beyond this, an even more innovative class of artificial neural network architectures emerges: deep learning models directly and originally designed on the IFN structure itself.

The foundation of this idea lies in a \textit{homological progression}, inspired by the hierarchical construction of simplicial complexes from lower-dimensional simplices to higher-dimensional ones. In this context, an IFN can be viewed as a topological structure starting with vertices (0-simplices), which connect to form edges (1-simplices), edges connect into triangles (2-simplices), triangles into tetrahedra (3-simplices), and so on. These higher-order simplices encapsulate increasingly complex interrelationships among the data’s features and provide the basis for what are referred to as Homological Neural Networks (HNNs) \cite{wang2023homological}.

While this progression is not homology in the strict mathematical sense, it is \textit{homological} in that it reflects the organization of relationships across multiple dimensions within a topological framework. Each step in this progression corresponds to a meaningful aggregation of relationships within the dataset, analogous to how simplicial homology studies the ways simplices combine to form topological spaces.

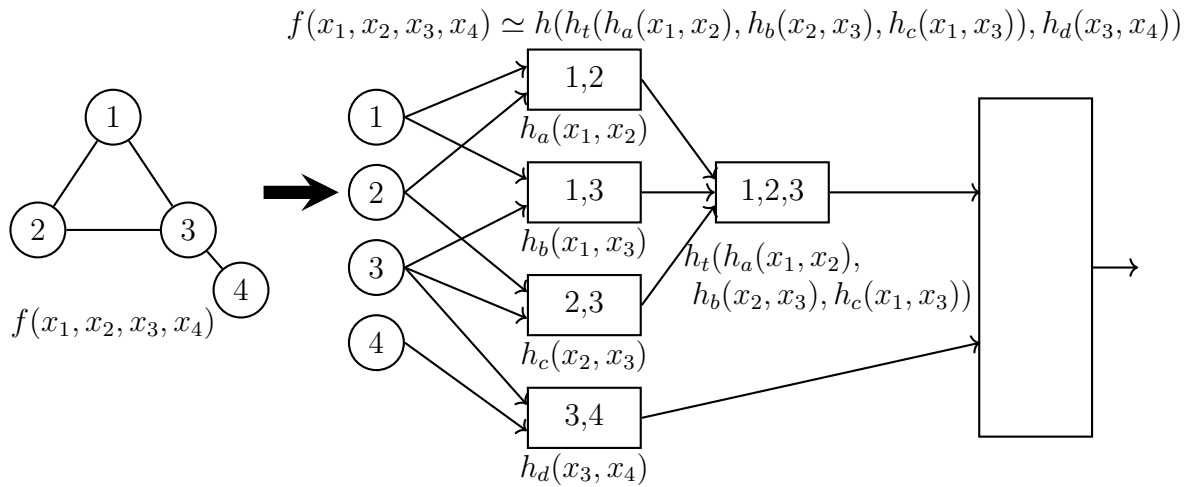
\begin{figure}
\begin{center}
\begin{tikzpicture}[node distance=1.5cm and 1.5cm, auto, thick]
    \node[circle, draw] (1l) at (-3,1) {1};
    \node[circle, draw] (2l) at (-4,-0.5) {2};
    \node[circle, draw] (3l) at (-2,-0.5) {3};
    \node[circle, draw] (4l) at (-1.3,-1.3) {4};

    \draw (1l) -- (2l) -- (3l) -- (1l) -- cycle;
    \draw (3l) -- (4l) ;
    
  \node[below=0cm of 2l, yshift=-0.5cm, xshift=1cm] (triangle_formula) 
    {$f(x_1, x_2, x_3, x_4)$};

\draw[->, line width=2.5pt, >=stealth, double distance=0mm] 
    (-1., 0) -- (-0, 0);
    
    \node[circle, draw] (1u) at (0.5,1) {1};
    \node[circle, draw] (2u) at (0.5,0) {2};
    \node[circle, draw] (3u) at (0.5,-1) {3};
    \node[circle, draw] (4u) at (0.5,-2) {4};

    \node[rectangle, draw, minimum width=1.5cm, minimum height=0.8cm, anchor=west] (12) at (2.5,1.5) {1,2};
    \node[rectangle, draw, minimum width=1.5cm, minimum height=0.8cm, anchor=west] (13) at (2.5,0) {1,3};
    \node[rectangle, draw, minimum width=1.5cm, minimum height=0.8cm, anchor=west] (23) at (2.5,-1.5) {2,3};
    \node[rectangle, draw, minimum width=1.5cm, minimum height=0.8cm, anchor=west] (34) at (2.5,-3) {3,4};

    \node[rectangle, draw, minimum width=1.5cm, minimum height=4.5cm, anchor=west] (last) at (8.5,-1) {};
   
    \node[below=0cm of 12, yshift=0.1cm] (formula_12) {$h_a(x_1, x_2)$};
    \node[below=0cm of 13, yshift=0.1cm] (formula_13) {$h_b(x_1, x_3)$};
    \node[below=0cm of 23, yshift=0.1cm] (formula_23) {$h_c(x_2, x_3)$};
    \node[below=0cm of 34, yshift=0.1cm] (formula_34) {$h_d(x_3, x_4)$};

    \node[rectangle, draw, minimum width=1.5cm, minimum height=0.8cm, anchor=west] (123) at (5.,0) {1,2,3};

    \draw[->] (1u.east) -- ([yshift=5pt]12.west);
    \draw[->] (2u.east) -- ([yshift=-5pt]12.west);

    \draw[->] (1u.east) -- ([yshift=5pt]13.west);
    \draw[->] (3u.east) -- ([yshift=-5pt]13.west);

    \draw[->] (2u.east) -- ([yshift=5pt]23.west);
    \draw[->] (3u.east) -- ([yshift=-5pt]23.west);

    \draw[->] (3u.east) -- ([yshift=5pt]34.west);
    \draw[->] (4u.east) -- ([yshift=-5pt]34.west);

    \draw[->] (12.east) -- ([yshift=5pt]123.west);
    \draw[->] (13.east) -- (123.west);
    \draw[->] (23.east) -- ([yshift=-5pt]123.west);

\node[below=0.1cm of 123,  xshift=.0cm, align=left] (formula1) 
    {$h_t\big(h_a(x_1, x_2),$};
\node[below=0.6cm of 123,  xshift=0.8cm, align=center] (formula2) 
     {$h_b(x_2, x_3), h_c(x_1, x_3)\big)$};

\draw[->] (34.east) -- ++(4.5,1);
\draw[->] (123.east) -- ++(2.,0);
\draw[->] (last.east) -- ++(0.6,0);

\node[above=0.6cm of last,  xshift=-4.0cm, align=left] (formula3) 
    {$f(x_1, x_2, x_3, x_4) \simeq h\big(h_t\big(h_a(x_1, x_2),h_b(x_2, x_3), h_c(x_1, x_3)\big),h_d(x_3, x_4)\big)$};
    
\end{tikzpicture}
\end{center}
\vskip-0.5cm
\caption{\label{f.HNN} 
An HNN layered deep architecture generated from an IFN made of a triangular clique and an attached edge. 
The HNN is approximating $f(x_1, x_2, x_3, x_4)$ as composite function $h\big(h_t\big(h_a(x_1, x_2),h_b(x_2, x_3), h_c(x_1, x_3)\big),h_d(x_3, x_4)\big)$ where the gathering of the variables in the IFN becomes a gathering into functions of functions. 
This approach to neural network architecture design was first proposed in \cite{wang2023homological}.}
\end{figure}

This homological progression can be directly mapped into a deep learning architecture, where vertices correspond to input nodes, and higher-order simplices (e.g., edges, triangles, or tetrahedra) represent intermediate or output layers. An illustrative example of such a layered HNN architecture, constructed from an IFN consisting of a triangular clique and an attached edge, is shown in Fig.~\ref{f.HNN}.
In this HNN architecture, the input layer comprises the three vertices of the triangle, and the vertex attached to it, the first hidden layer corresponds to the edges connecting these vertices, and the edge between vertex 4 and the triangle. The second hidden layer represents the triangle as a higher-order (2D) simplex. Instead, the node associated with the edge between node 4 and the triangle outputs directly into the final aggregation layer. Each node in the layers corresponds to a distinct aggregation of the input variables. For instance, the node \((1,2) \) processes variables \(x_1\) and \(x_2\) using a function \(h_a(x_1, x_2)\). The overall HNN model can be expressed as a composite function 
\(
f(x_1, x_2, x_3, x_4) \simeq h\big(h_t\big(h_a(x_1, x_2),h_b(x_2, x_3), h_c(x_1, x_3)\big),h_d(x_3, x_4)\big).
\)
At this level of abstraction, the nature of the units and functions is flexible and can be tailored to specific tasks. In \cite{wang2023homological}, the system used multilayer perceptrons (MLPs), and the aggregation functions were implemented as ReLU activations applied to weighted sums of the variables. For instance: 
\(
h_a(x_1, x_2) = \text{ReLU}(w_1 x_1 + w_2 x_2 + b),
\)
where \(w_1, w_2\) are the learnable weights, and \(b\) is the bias term. 
Alternatively, this architecture could be set as a sparse Kolmogorov-Arnold network (KAN). In such a case, the aggregation functions could take the  form:
\(
h_a(x_1, x_2) = \Phi_{1,1}(x_1) + \Phi_{1,2}(x_2) 
\).
Additionally, the HNN architecture can be extended with readout elements that extract information from each layer, followed by a final aggregation unit, enabling flexible modeling and integration of multiscale dependencies.

\begin{figure}
\begin{center}
\begin{tikzpicture}[scale=1,node distance=1.5cm and 1.5cm, auto, thick]

    \node[circle, draw] (1l) at (-6.3,1) {1};
    \node[circle, draw] (2l) at (-7.3,-0.5) {2};
    \node[circle, draw] (3l) at (-5.3,-0.5) {3};
    \draw (1l) -- (2l) -- (3l) -- (1l) -- cycle;

    \node[below=0cm of 2l, yshift=-0.5cm, xshift=1cm] (triangle_formula) 
        {$f(x_1, x_2, x_3)$};

\draw[->, line width=2.5pt, >=stealth, double distance=0mm] 
    (-4.8, 0) -- (-4, 0);

    \node[circle, draw] (1u) at (-3.5,1.5) {1};
    \node[circle, draw] (2u) at (-3.5,0) {2};
    \node[circle, draw] (3u) at (-3.5,-1.5) {3};

    \node[rectangle, draw, minimum width=1.5cm, minimum height=0.8cm, anchor=west] (12) at (-1,2) {1,2};
    \node[rectangle, draw, minimum width=1.5cm, minimum height=0.8cm, anchor=west] (13) at (-1,0.5) {1,3};
    \node[rectangle, draw, minimum width=1.5cm, minimum height=0.8cm, anchor=west] (23) at (-1,-1) {2,3};
    \node[rectangle, draw, minimum width=1.5cm, minimum height=0.8cm, anchor=west] (123) at (-1,-2.5) {1,2,3};

    \node[below=0cm of 12, yshift=0.1cm] (formula_12) {$h_a(x_1, x_2)$};
    \node[below=0cm of 13, yshift=0.1cm] (formula_13) {$h_b(x_1, x_3)$};
    \node[below=0cm of 23, yshift=0.1cm] (formula_23) {$h_c(x_2, x_3)$};
    \node[below=0cm of 123, yshift=0.1cm] (formula_123) {$h_d(x_1, x_2,x_3)$};

    \node[rectangle, draw, minimum width=3cm, minimum height=4.5cm, anchor=west] (final) at (3,0) { };

    \draw[->] (1u.east) -- ([yshift=5pt]12.west);
    \draw[->] (2u.east) -- ([yshift=-5pt]12.west);

    \draw[->] (1u.east) -- ([yshift=5pt]13.west);
    \draw[->] (3u.east) -- ([yshift=-5pt]13.west);

    \draw[->] (2u.east) -- ([yshift=5pt]23.west);
    \draw[->] (3u.east) -- ([yshift=-5pt]23.west);

    \draw[->] (1u.east) -- ([yshift=5pt]123.west);
    \draw[->] (2u.east) -- (123.west);
    \draw[->] (3u.east) -- ([yshift=-5pt]123.west);

    \draw[->] (12.east) -- ([yshift=1.5cm]final.west);
    \draw[->] (13.east) -- ([yshift=0.5cm]final.west);
    \draw[->] (23.east) -- ([yshift=-0.5cm]final.west);
    \draw[->] (123.east) -- ([yshift=-1.5cm]final.west);

    \node[below=-.5cm of final, yshift=-0.5cm, xshift=-1.cm] (formula_output) 
     {$h\big(h_a(x_1, x_2), h_b(x_2, x_3),$};
   \node[below=0cm of final, yshift=-0.5cm, xshift=-0.5cm] (formula_output) 
        {$h_c(x_1, x_3), h_d(x_1, x_2,x_3)\big)$};

\end{tikzpicture}
\end{center}
\vskip-0.5cm
\caption{\label{f.HCNN} Another simple HNN architecture generated from a IFN made of a single triangular clique. 
This architecture was introduced in \cite{briola2023homological} with convolutional units and given the name of HCNN.}
\end{figure}
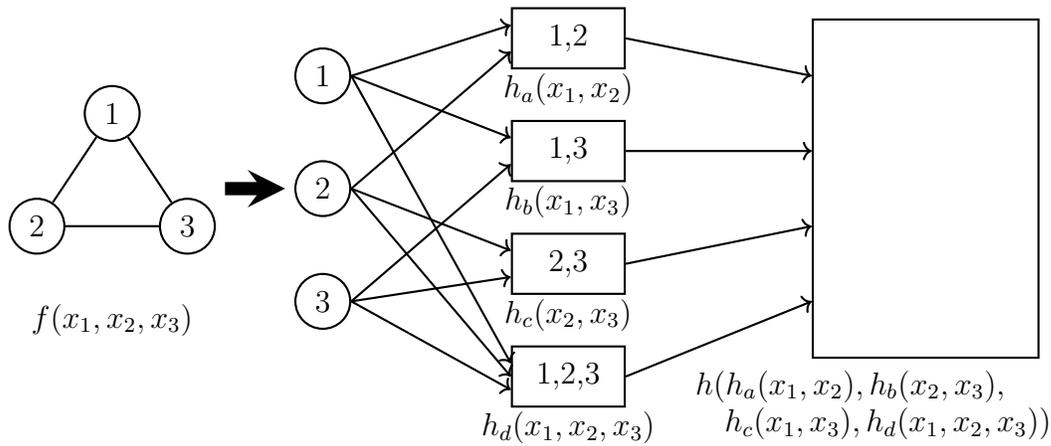

Another illustrative example of an HNN architecture,  derived from an IFN consisting of a single triangular clique, is shown in Fig.~\ref{f.HCNN}. Unlike the previous architecture, all homological elements in this design are present in the first hidden layer. This configuration, termed homological convolutional neural network (HCNN), was introduced in \cite{briola2023homological}, where the first layer’s operations involved convolutions applied across the simplicial elements.
In this architecture, information is aggregated in convolutional nodes of increasing dimensionality, enabling the neural network to learn models shaped by the dependency structure encoded in the IFN. These convolutions mimic the local operations performed in geometrical spaces—such as sliding-window 2D convolutions in image processing—but rely on topological distance and depth rather than geometric proximity, thus enabling high-dimensional convolution.
The HCNN serves as a modular unit that can be seamlessly connected with other pre-processing or post-processing components. For example, in \cite{briola2023homological}, a Long Short-Term Memory (LSTM) network was employed as a pre-processing layer to refine input features before passing them into the HCNN that was then feeding a multi-layer perceptron.

\section{Applications of IFNs }\label{s.Applications}
Initially introduced for modeling \emph{financial systems} as complex, interrelated structures \cite{mantegna1999,tumminello2005}, IFNs have primarily been applied within this domain. However, under the broader framework of complex systems modeling, recent years have witnessed a rapid expansion of their applications across a much wider range of fields.

\begin{figure}
 \begin{center}
\includegraphics[width=0.85\textwidth]{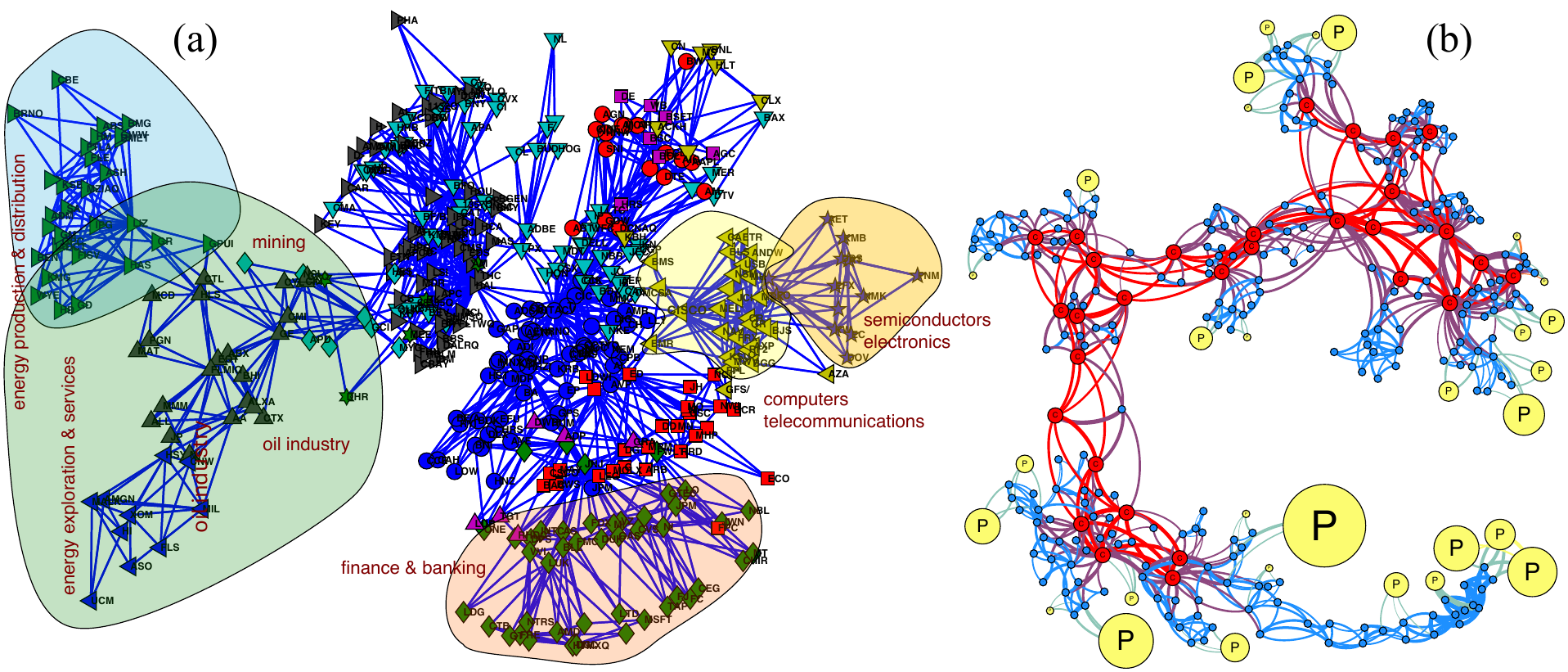}
 \end{center}
 \caption{\label{f.PMFGclustering} 
 (a) An example of structural organization and clustering from the PMFG structure from correlations between log-returns of 400 qualities of the S\&P500 observed over the period between 1996-2009 (see \cite{musmeci2015relation} for details). 
 (b) An example of portfolio selection from a PMFG network based on centrality position. This PMFG network is realized from correlations of  300 us stocks during the period 1981-2010 (see \cite{pozzi2013spread} for details). 
}
\end{figure}

\subsection{ IFNs for Finance: Market Structure, Spread of Financial Risk, Portfolio Selection and Investment Differentiation }
IFNs, such as the MST, PMFG, TMFG, and MFCF, have been extensively employed to analyze dependencies between financial equities, uncovering systemic risks, and revealing hierarchical structures in financial markets \cite{chi2010network,bardoscia2021physics}. These tools have proven instrumental in identifying clusters of correlated assets, providing actionable insights for risk management, enhancing the understanding of market dynamics during crises, and contributing to the protection of investors through better-informed decision-making \cite{saha2022survey}. 
As an example, from \cite{musmeci2015relation}, Fig.~\ref{f.PMFGclustering}(a) shows an example of market structure filtered by the PMFG from 400 qualities of the S\&P500 over the period between 1996-2009. One can see that industry activities are captured within the structure of the IFN with, for instance, energy, mining, and oil and gas industries gathered on the left side, while semiconductors and technology are on the right side of the PMFG.  
IFNs have been applied to study the hierarchical structure of interest rates \cite{di2005interest} and the organization of financial markets across industries and economic sectors, often extending beyond traditional classifications \cite{di2010use, musmeci2015relation, musmeci2017multiplex}. They have also provided critical insights into the distribution of risk \cite{di2005interest} and the complex interplay between market structures and financial instability \cite{musmeci2016interplay}, including in decentralized financial systems \cite{briola2023anatomy}. By revealing how heterogeneous risk spreads across financial markets relate to the structure of IFNs, these methods have supported the development of effective diversification strategies \cite{pozzi2013spread, musmeci2014risk} and clarified the effects of financial crises and instabilities on market structures \cite{aste2010correlation, procacci2020market, aste2021stress, seabrook2022quantifying}. They have also played a vital role in guiding portfolio optimization by leveraging sector-specific dependencies \cite{turiel2020sector} and topological insights \cite{procacci2022portfolio, wang2023dynamic, wang2023topological}, while supporting dynamic investment strategies through the analysis of the evolving structure of financial networks \cite{musmeci2014risk}. In the context of cryptocurrency markets, IFNs have been applied to uncover the emergent structure of these markets and to explore the interrelation between market structures and sentiment changes in investors and citizens \cite{aste2019cryptocurrency, briola2022dependency} including in the emerging ESG and `green' financial eco-systems \cite{stoccoESG24}. They have also been used to predict structural relations between sentiment and price dynamics \cite{souza2019predicting}, offering tools to better understand the drivers of market behavior and the mechanisms that contribute to the formation of market structures. Furthermore, the persistent dynamical properties of IFNs have made them excellent instruments for defining and forecasting market states and their evolution \cite{turiel2022simplicial, procacci2021forecasting}, enabling proactive strategies to mitigate risks and foster market stability. 

Figure~\ref{f.PMFGclustering}(b), from \cite{pozzi2013spread}, shows how stocks in a portfolio can be selected based on their centrality position in the PMFG network. For this example, it turns out that a selection of the most peripheral stocks (yellow circles) is better at managing risks and performance than a selection of the most central ones (red circles). Several studies and a large number of practical applications have supported the finding that risk does not distribute uniformly in IFNs and, in particular, the center and periphery have very different risk profiles. Where is better to invest, periphery or center, depends on the system and the time period, with investments in peripheral assets tending to be more resilient to market stress \cite{turiel2020sector,procacci2022portfolio}.  
Similar outcomes were found in \cite{wang2023topological} where a Statistically Robust version of IFN (SR-IFN) is used to sparsify asset correlations and identify peripheral assets that enhance diversification and reduce risk. The methodology leverages network centrality measures to guide portfolio selection and weighting, achieving significant improvements in risk-adjusted returns across various market indices.

\subsubsection{ IFN-LoGo Covariance Filtering for Portfolio Optimization }

Portfolio optimization involves methodologies designed to select and weight equities to achieve specific financial objectives, such as maximizing positive returns while minimizing negative fluctuations in the portfolio's returns. The field has undergone significant advancements, with a variety of established and novel methods emerging, increasingly incorporating machine learning and deep learning techniques to enhance optimization outcomes.
All these approaches fundamentally rely on leveraging the dependency structure of the variables, combining assets in a manner that optimally distributes risk to reduce (negative) variability while maintaining good returns. At the foundation of portfolio optimization lies Markowitz’s modern portfolio theory \cite{Markowitz52,markowitz1968portfolio}, which provides an optimal solution for a multivariate set of returns under the assumption that expected values and covariances are well-defined and known. A central element of this theory is the inverse covariance matrix, which plays a crucial role in determining the optimal portfolio weights. Specifically, the weights for the portfolio with a desired return and minimal variance are given by: 
\(\mathbf{w} =  \boldsymbol{\Sigma^{-1}} \left(\lambda  \boldsymbol{\mu} + \gamma  \mathbf{1} \right) \), where \(\lambda\) and \(\gamma\) are two scalar parameters (Lagrange multiplyers), \( \boldsymbol{\mu} \) is the vector of expected values, \( \mathbf{1} \) a column vector of ones, and \( \boldsymbol{\Sigma^{-1}} \) is the inverse covariance \cite{Markowitz52,markowitz1968portfolio}. 
The appeal of Markowitz’s formula for portfolio weights is that it is the exact optimal solution. 
Other methods often have less explicit dependence on the covariance matrix, yet they all rely critically on an accurate estimation of the dependency structure. A major challenge for all these methods lies in the fact that the optimization must be performed for future returns, requiring both the estimations of the expected values and the dependency structure to align with future market conditions. This is where IFN-LoGo provides a significant advantage: by filtering information from past observations, IFNs extract a filtered dependency structure that is less affected by the curse of dimensionality, more robust to noise, and more likely to remain relevant in the future. For instance, the application of IFNs to the Markowitz’s optimization can be trivially done by substituting the inverse covariance with its sparse estimate: \( \boldsymbol{\Sigma^{-1}} \leftarrow \mathbf{J}^{sp}\). Other refinements such as considering the centrality of the stocks in the IFN as a selection criterion or looking at persistent structures in the dynamics of IFNs constructed over rolling windows can further improve performance and robustness \cite{pozzi2013spread,turiel2020sector,procacci2022portfolio}. 

The advantage of employing IFN-LoGo sparse inverse covariance has been demonstrated in several studies, both for simple implementations of Markowitz portfolios and for more sophisticated approaches, such as  Least-Square Error (LSE),  or Black-Litterman models  \cite{procacci2022portfolio,wang2023dynamic,wang2023topological}. By simply replacing the full covariance matrix with the sparse inverse covariance derived from IFN-LoGo, these methods achieve superior portfolio performance without requiring any new methodology. Instead, they leverage the improved precision and interpretability of IFNs to enhance existing optimization frameworks. The resulting portfolios consistently outperform those constructed using the full covariance matrix, underscoring the practical benefits of incorporating IFNs into portfolio optimization. Moreover, the sparse dependency structure extracted by IFNs offers deeper insights into the dynamics of the system under analysis. As a dynamical property, this structure fluctuates stochastically over time, reflecting changes in the system’s state. By leveraging the filtering capabilities of IFNs, these structural dynamics can be identified, tracked, and even forecasted, enabling a more nuanced understanding of market conditions and enhancing predictive modeling \cite{procacci2021forecasting}.

\begin{figure}
 \begin{center}
\includegraphics[width=0.65\textwidth]{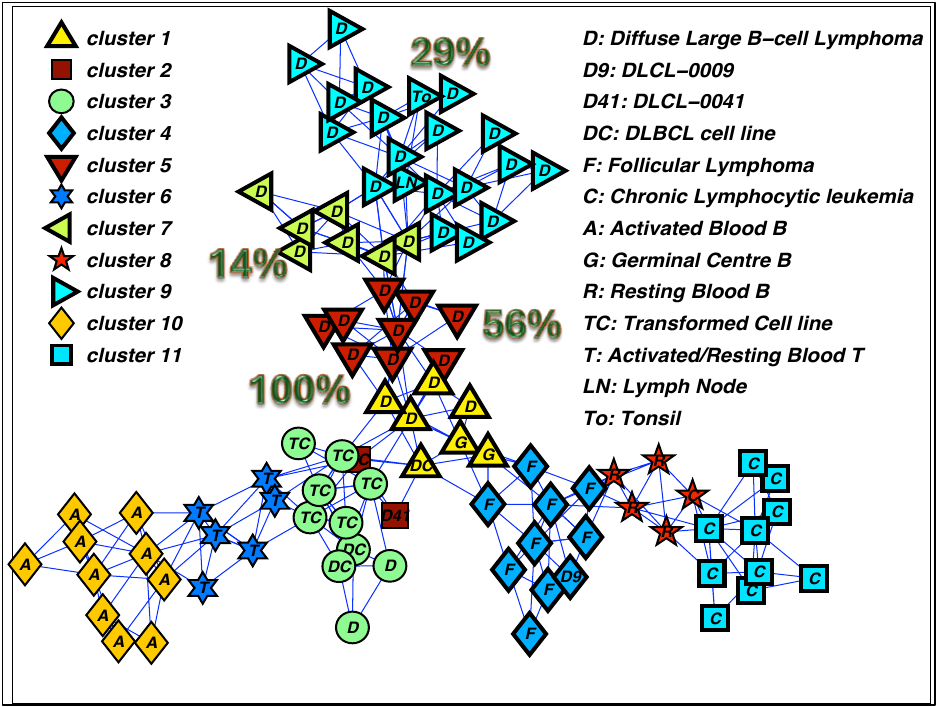}
 \end{center}
 \caption{\label{f.Cancer} 
PMFG network and cluster structure for 96 malignant and normal lymphocyte samples from \cite{alizadeh2000distinct}. The labels inside the symbols indicate the sample types, as detailed in the legend. The DBHT technique identifies 11 distinct sample clusters, represented by different symbols. Notably, in the IFN structure, the Diffuse Large B-cell lymphoma is divided into four clusters (1, 5, 7, 9), each associated with markedly different five-year survival rates (percentages reported on the side of the clusters).  
}
\end{figure}

\subsection{ IFNs Hierarchical Clustering for Biology }

The structure of IFNs provides a natural framework for data clustering. As highlighted in \cite{song2011nested}, the IFN structure is organized in a nested hierarchical manner, where separators act as subsets dividing the vertex set into an ``inside'' and ``outside,'' thereby establishing a partial order set (poset). This partial order, embedded in the clique tree structure, defines a unique hierarchy that can be directly leveraged for clustering. Such hierarchical clustering, as demonstrated in \cite{song2012hierarchical} for the PMFG case, enables the deterministic extraction of clusters and hierarchies in complex datasets.  Being posets implies that Hasse diagrams can be employed to represent IFNs directly. The poset properties of simplicial complexes have recently garnered attention in topological data analysis \cite{billings2019simplex2vec}, highlighting the broader applicability and utility of IFNs.

In the field of biology, hierarchical clustering based on IFNs has been successfully applied to gene expression data for identifying regulatory modules and interactions \cite{song2015multiscale,xu2022landscape}. In particular, \cite{song2012hierarchical} introduced an IFN-based approach to extract biologically meaningful clusters in an unsupervised and deterministic manner without requiring prior assumptions or information. Applying this method to gene expression patterns of lymphoma samples, they uncovered significant groups of genes that play critical roles in the diagnosis, prognosis, and treatment of major human lymphoid malignancies.
The paper \cite{song2012hierarchical} underscored the relevance of triangular motifs in gene expression and, more broadly, the significance of IFN architectures that are locally dense yet globally sparse. This application of IFNs to gene expression laid the groundwork for subsequent graph-based methodologies, such as \cite{hu2021hiscf}, which also emphasize the role of triangular motifs in gene expression.
Figure \ref{f.Cancer} illustrates one of the key findings, showcasing IFN-based clustering (referred to as DBHT in the original work) applied to data from 96 malignant and normal lymphocyte samples derived from \cite{alizadeh2000distinct}. The IFN structure effectively partitions the dataset into distinct clusters. Some clusters align with well-known lymphocyte families, such as clusters 10 and 11, corresponding to Follicular Lymphoma and Chronic Lymphocytic Leukemia, respectively. Others, like the Diffuse Large B-cell lymphoma, are divided into four distinct clusters (1, 5, 7, and 9).
Notably, these clusters reveal significant variations in patient outcomes, as evidenced by the five-year survival rates: cluster 1 shows a survival rate of 100\%, cluster 5 has 56\%, cluster 7 has 14\%, and cluster 9 has 29\%. This striking example underscores the power of IFN-based hierarchical clustering to uncover clinically and biologically relevant patterns in complex datasets. It highlights its potential for providing deeper insights into disease heterogeneity, guiding prognosis, and informing therapeutic strategies.

\subsection{IFNs for Climate Science and Social Sciences}
In {climate science}, IFNs have been used to analyze global climate networks by filtering significant correlations between climate indices, allowing researchers to identify regional climate interactions and study the propagation of climatic events \cite{donges2009,andjelkovic2024rainfall}. These methods help isolate critical dependencies while mitigating the influence of spurious correlations.

The {social sciences} have also benefited from IFN applications, particularly in the analysis of social networks and sentiment propagation. By filtering noise from raw data, IFNs have been used to identify influential nodes and communities, as well as to study information spread in large-scale social networks \cite{barabasi1999,goh2014complex,dai2017temporal,golino2022modeling}.

\begin{figure}
 \begin{center}
\includegraphics[width=0.65\textwidth]{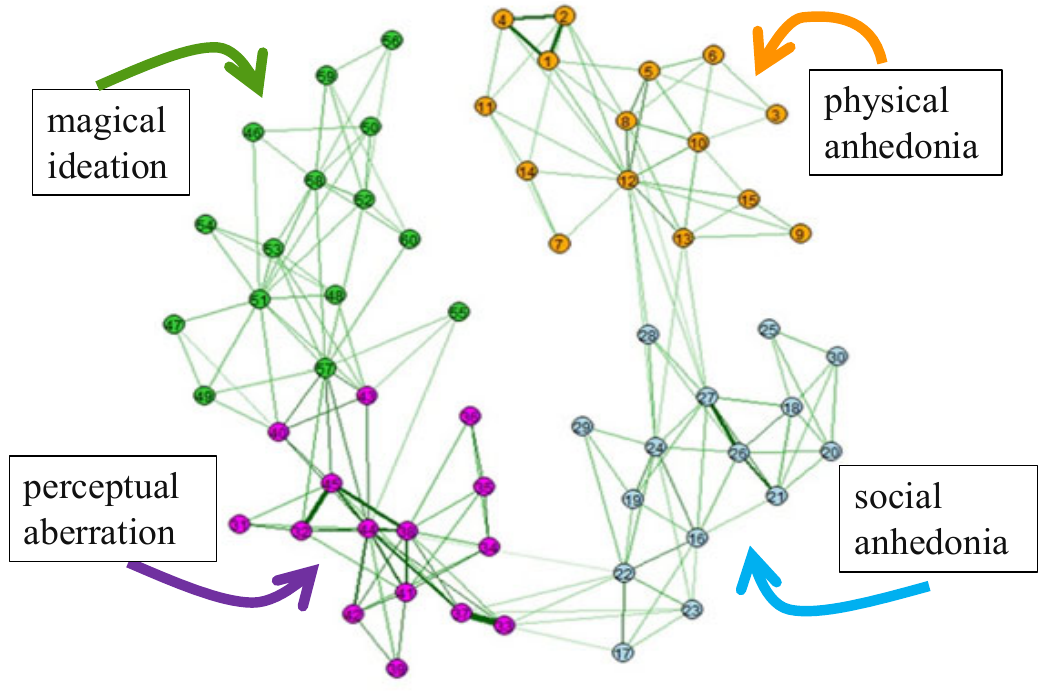}
 \end{center}
 \caption{\label{f.Psyco} 
TMFG structure from \( 60 \times 60 \) WSS-SF endorsement association matrix computed form a psycometric survey of 5,831
participants (see \cite{christensen2018network}). 
}
\end{figure}

\subsection{IFNs for Psychology and Neurosciences}

In psychology, IFNs have proven to be powerful tools for analyzing complex relationships within psychological constructs and behavioral data. By filtering noisy correlations and retaining only the most significant interactions, IFNs provide a robust framework for psychometric analyses and psycology studies \cite{christensen2018network,alexander2010disrupted,benedek2017semantic,christensen2018,christensen2024comparing,christensen2021estimating,coburn2020psychological,kenett2014investigating,rastelli2022simulated,chen2023mapping,kenett2023aesthetic,kaiser2024increasing}. 

A notable and pioneering application of IFNs in the psychological domain is found in the study of schizotypy—a construct associated with schizophrenia-spectrum disorders. In \cite{christensen2018}, the Wisconsin Schizotypy Scales–Short Forms (WSS-SF), a 60-item true–false questionnaire, was employed to assess positive and negative schizotypy in a sample of 5,831 participants. Positive schizotypy was measured using subscales for perceptual aberration (distortions of body image) and magical ideation (delusional or irrational beliefs), while negative schizotypy was evaluated through subscales for physical anhedonia (reduced sensory pleasure) and social anhedonia (lack of interest in social interactions).
An endorsement association matrix was constructed from Pearson’s correlations, quantifying the likelihood of endorsing one item given the endorsement of another. The TMFG method was then applied to construct an IFN from this matrix, efficiently capturing the most relevant interactions between items while minimizing spurious associations. The resulting TMFG structure, shown in Fig.~\ref{f.Psyco}, clearly distinguishes between the positive and negative schizotypy subscales, demonstrating the network’s capacity to capture meaningful psychological constructs. 
A key advantage of the IFN approach is its ability to generate hierarchical structures that align with the dimensional nature of many psychological and psychopathological phenomena. In the context of the WSS-SF, local connections between individual items naturally aggregate into symptom clusters, which in turn map onto broader scales or phenomena. This hierarchical organization provides a nuanced understanding of overlapping symptom nuances and their contributions to larger symptom domains, such as those associated with schizophrenia-spectrum disorders.
The IFN-based networks also demonstrated clinical relevance by producing connections that were consistent with traditional findings in schizophrenia-spectrum liability. Furthermore, the study highlighted the reliability and theoretical consistency of IFNs in reproducing both global and local network characteristics across samples. Importantly, IFNs exhibited strong predictive validity for positive and negative schizotypy, outperforming lasso-based networks in terms of overall predictability. These findings underscore the potential of IFNs as a reliable network-based approach in psychometric analysis, contributing to the ongoing discussion on reproducibility and predictive accuracy in psychological research \cite{christensen2024comparing,christensen2021estimating,rastelli2022simulated}.

In the adjacent domain of {neuroscience}, IFNs have been applied to functional brain networks to reveal meaningful topological structures. These methods are particularly useful in understanding brain connectivity, as they can uncover patterns of activity while minimizing the effect of indirect correlations, thus enhancing interpretability \cite{alexander2010disrupted,de2017topological,christensen2018networktoolbox,jiang2021toward,borne2024unveiling}

\subsection{IFNs for Artificial Intelligence and Deep Learning}

Beyond human intelligence and the human brain, IFNs have demonstrated significant potential in artificial intelligence, particularly in deep learning applications. As discussed in Section \ref{s.QuantitativeG}, IFNs provide a natural framework for integrating network structures with quantitative modeling approaches. This integration facilitates tasks such as feature selection and the development of specialized deep-learning architectures. In the following subsections, I will highlight some key findngs and advancements in this area.

\subsubsection{IFNs for Feature Selection for Tabular Data Classification}\label{s.TFS}

As introduced in Section \ref{ss.Feature}, one practical application of IFNs is in feature selection for high-dimensional datasets. 
A particularly effective method, based on IFNs, is the TFS (see Section \ref{s.TFS} and \cite{briola2023topological}), 
which, in \cite{briola2023topological}, was benchmarked against other state-of-the-art methods, such as Infinite Feature Selection (Inf-FS), on 16 diverse datasets spanning application domains like text, image, biological, and artificial data. Results show that TFS consistently matches or outperforms Inf-FS in terms of accuracy, stability, and computational efficiency, particularly for tabular data. 
 Notably, TFS is advantageous for handling both linear and non-linear relationships between features, with options to customize the similarity metrics used to construct the TMFG, such as Pearson, Spearman,  Energy coefficients, or any similarity measure. This flexibility allows TFS to adapt to a wide range of data characteristics, making it a robust tool.

\subsubsection{Combining IFNs and Graphical Neural Networks for Classification and Time Series Analysis Tasks }

In \cite{wang2022network}, a spatial-temporal GNN architecture, the Filtered Sparse Spatial-Temporal GNN (FSST-GNN), was employed to predict future sales volumes from a Kaggle dataset of 5-year sales data across 50 products in 10 stores. The IFN-generated sparse networks from MFCF provided the structural input for the GNN’s spatial component, enabling superior performance compared to fully connected, disconnected, and unfiltered graph structures. 
The study demonstrated that FSST-GNN outperformed benchmark models, including the Diffusion Convolutional Recurrent Neural Network (DCRNN), GNN-ARMA, and LSTM without graphical information. While the graphical LASSO (GLASSO) filtering method achieved the best overall performance, IFN-based filtering offered nearly equivalent results with three times lower computational complexity, greater interpretability, and a simpler architecture. Moreover, the sparsity inherent to IFNs enabled efficient attention computation in graph attention networks (GATs) and improved graph convolutional networks (GCNs) by providing interpretable weighted adjacency matrices.
This integration illustrates that, even with a relatively basic GNN architecture, the use of advanced filtering methods like IFNs can significantly enhance model accuracy, scalability, and interpretability in time-series forecasting tasks.

\subsubsection{ HNN for regression, classification, and forecasting}

HNNs (see Section \ref{ss.HNN}) are highly effective in traditionally challenging domains for deep learning, such as tabular data and multivariate time-series regression. For instance, in \cite{wang2023homological} we applied the HNN architecture (as the one sketched in Fig.\ref{f.HNN}) on the Penn Machine Learning Benchmark (PMLB) and multivariate time-series datasets. Experiments show that HNN consistently outperformed standard and sparse Multi-Layer Perceptrons (MLPs) with the same depth and number of neurons. The results demonstrated that the sparse, higher-order homological structure of HNN, combined with residual connections and readout units, plays a key role in enhancing performance.

Also in \cite{wang2023homological}  it is shown that, when extended to temporal modeling (e.g., LSTM-HNN), HNNs delivered superior predictive accuracy in time-series forecasting compared to baseline MLPs, sparse MLPs, and other advanced models like RNN-GRU and LSTNet. While slightly outperformed by state-of-the-art models such as MTGNN and TPA-LSTM, HNNs achieved comparable results with a fraction of the parameters, demonstrating greater computational efficiency and interpretability. Notably, HNNs reduced error metrics (e.g., RSE) by significant margins across various forecasting horizons and datasets.

HNNs stand out for their ability to align performance with state-of-the-art models while maintaining simplicity and reducing parameter count. Their hierarchical structure eases the computational process, while the sparse design enhances scalability and interpretability, offering a robust alternative to more complex, parameter-heavy models. These findings position HNNs as an efficient, interpretable, and powerful framework for data-driven modeling in high-dimensional and time-series contexts.

\subsubsection{ HCNN for Tabular Data Classification and the Limit Order Book Forecasting }

A notable implementation of HCNN  (see Section \ref{ss.HNN} and Fig.~\ref{f.HCNN}), introduced in \cite{wang2023homological}, is BootstrapNet, which incorporates a bootstrapping procedure over IFNs to enhance robustness to data noise and improve the representation of dependency structures. This method generates multiple replicas of the original dataset by randomly sampling rows with replacement. For each replica, a TMFG is constructed, and the final graph is obtained by retaining only those edges that appear with a frequency above a predefined threshold across the replicas. The result is a sparse and noise-resistant network representation, well-adapted to the structure of the data and significantly more computationally efficient than traditional approaches.
Experimental results, reported in \cite{wang2023homological}, demonstrate HCNN’s superior performance across various tabular numerical data classification tasks. The BootstrapNet configuration, in particular, excels in balancing computational efficiency and prediction accuracy. HCNN consistently outperforms classic machine learning models, such as Logistic Regression, Random Forest, and XGBoost, and matches or exceeds the performance of state-of-the-art deep learning models like TabPFN and TabNet. Importantly, compared to other attention-based architectures, HCNN achieves comparable results with significantly fewer parameters and a more interpretable structure, making it a compelling choice for tasks requiring both scalability and robust modeling.

Another application of the HCNN architecture is in mid-price change forecasting from Limit Order Book (LOB) data, as presented in \cite{briola2024hlob}. By applying convolutions over cliques and sub-cliques derived from dependency structures among LOB volume levels, the model effectively captures the spatial organization of information while ensuring high computational efficiency and interpretability. A key feature of the architecture is the use of a bootstrapping procedure in the computation of mutual information matrices, which enhances the robustness of the structural priors.
The HLOB architecture was evaluated on three datasets comprising 15 NASDAQ-listed stocks, grouped by tick size (small, medium, and large). Experimental results showed that HLOB outperformed nine advanced deep learning models—including transformer-based and hybrid CNN-transformer architectures—in 73.3\% of scenarios. Notably, for medium- and large-tick stocks, HLOB consistently achieved top F1 scores (0.41 for medium-tick, 0.48 for large-tick) and strong Matthews Correlation Coefficient (MCC) values (0.16 and 0.33, respectively). These results highlight the model’s effectiveness, particularly in environments with stable, hierarchical spatial structures, such as those characteristic of large-tick stocks.

In \cite{briola2024hlob}, the integration of bootstrapping into the mutual information computation process ensures robust network representations, which form the core of the TMFG-based structural priors used in HLOB. This approach enhances the model’s adaptability, allowing HLOB to perform particularly well in short-term prediction horizons for small- and medium-tick stocks, where informational drift is more pronounced. Moreover, the architecture is well-suited as a foundational tool for analyzing the hierarchical and dynamically evolving spatial dependency structures inherent in LOB data.

Future directions for advancing HCNN-based models, as outlined in \cite{briola2024hlob}, include the refinement of similarity metrics to accommodate mixed data types and the extension of the architecture to incorporate temporally evolving IFNs. These developments would allow the models to dynamically adapt to shifts in financial market microstructures, further strengthening the integration between deep learning and microstructural modeling.

\section{Conclusions} \label{s.Conclusions}

This review has examined the theoretical, algorithmic, and practical foundations of Information Filtering Networks (IFNs), underscoring their relevance for modeling complex systems through sparse, interpretable structures. By capturing the underlying dependency architecture in high-dimensional data, IFNs serve as effective tools for filtering noise, revealing structure, reducing data dimensionality and complexity, and supporting quantitative modeling.

A central contribution of IFNs lies in their ability to balance global sparsity with local density, resulting in network representations that are both scalable and rich in informational content. Efficient generative algorithms such as TMFG and MFCF (see Section~\ref{s.GenerativeIFNalgos}) enable the construction of these networks in polynomial time, offering practical solutions to the otherwise intractable problem of capturing multivariate conditional dependencies.

Crucially, IFNs bridge structural modeling with functional representation. Their alignment with composable functions provides a principled mechanism for integrating dependency structures into statistical and machine-learning models. This is particularly impactful in graphical modeling, where IFNs support the estimation of sparse inverse covariance matrices that rival or outperform traditional methods like Graphical LASSO—offering superior interpretability and reduced computational complexity. In domains such as portfolio optimization, IFNs make the modeling pipeline more transparent by linking risk-based dependency structures directly to optimization-ready covariance matrices.

The versatility of IFNs has been demonstrated across numerous disciplines  (see Section~\ref{s.Applications}). In finance, they underpin innovations in portfolio construction and high-frequency trading systems. In biology, they enhance the interpretability of gene expression analyses. In psychology and neuroscience, they support the mapping of cognitive and behavioral dependencies. In artificial intelligence, IFNs contribute to feature selection, explainable modeling, and the design of novel architectures such as HNNs and HCNNs.

Looking forward, the integration of IFNs into machine learning and deep learning frameworks presents fertile ground for research. Embedding structured priors into models not only improves interpretability and computational efficiency but also mitigates challenges such as overfitting and noise sensitivity. Moreover, IFNs provide a natural mechanism for encoding spatial and temporal attention—via dependency structures and their lagged counterparts—suggesting a promising role in the design of next-generation learning architectures.

Several directions merit further investigation. These include refining similarity and gain functions, scaling IFNs to ultra-high-dimensional settings, and incorporating dynamics through state-dependent or temporally evolving networks. Importantly, IFNs offer a form of topological embedding that informs models about the structural organization of data, potentially improving generalization and robustness.

While IFNs have seen successful adoption in fields like finance and biology, other domains remain underexplored. In particular, natural language processing and image analysis represent compelling frontiers where IFNs could yield substantial insights. With the exception of early applications in NLP \cite{gramatica2014graph}, there is little precedent—but significant potential—for IFN methodologies in these areas.

In summary, IFNs offer a principled, flexible, and interpretable framework for representing complex dependencies in data and exploiting them to improve data-driven modeling. As both their theoretical foundations and computational tools continue to evolve, IFNs are poised to play an increasingly central role at the intersection of network science, machine learning, and real-world applications.

Looking further ahead, the advent of quantum computing may radically reshape the computational landscape underpinning IFNs. To date, only the MST admits efficient exact solutions. For all other IFN-related problems, one must necessarily trade off optimality for tractability, yielding suboptimal solutions within polynomial time. However, quantum computing offers the potential to transcend these classical limits. If realized, this could revolutionize how IFNs are constructed and optimized, enabling exact or near-optimal solutions to problems that are currently only heuristically addressed. As quantum hardware and algorithms mature, their integration with IFN frameworks may unlock a new era of scalable, high-fidelity, and interpretable modeling in complex systems.

\section*{Acknowledgments}
I would like to express my greatest gratitude to the entire Financial Computing and Analytics group at UCL, Computer Science, for their invaluable support. Special thanks go to all my former and present PhD students, who greatly contributed to shaping my research. Let me mention in particular:  Wolfram Barfuss, Antonio Briola, Hongyu Lin, Guido Massara, Pier Francesco Procacci, Marwin Schmidt, Yo-Der Song, Jeremy Turiel, and Raymond Wang for their dedication and huge contributions. 

This work was not funded by any specific grant or external support apart from my salary from UCL. However, it represents over two decades of research, much of which was made possible through numerous grants. Since acknowledging these grants contributes to their impact metrics, I take this opportunity to do so: ARC, Australia DP0450292; APAC Australia; SRC, United Kingdom (ES/K002309/1); EPSRC, United Kingdom (EP/P031730/1); EC (H2020-ICT-2018-2 825215).

\bibliographystyle{unsrt} 

\end{document}